\documentclass{article}

\usepackage{microtype}
\usepackage{graphicx}
\usepackage{subfigure}
\usepackage{booktabs} % for professional tables

\usepackage{hyperref}

\usepackage[accepted]{icml2024}

\usepackage{amsmath}
\usepackage{amssymb}
\usepackage{mathtools}
\usepackage{amsthm}

\usepackage[capitalize,noabbrev]{cleveref}

\theoremstyle{plain}

\theoremstyle{definition}

\theoremstyle{remark}

\usepackage[textsize=tiny]{todonotes}

\icmltitlerunning{\textit{ETHER}: Efficient Finetuning of Large-Scale Models with Hyperplane Reflections}

\usepackage{soul}
\usepackage{color}
\usepackage{float}
\usepackage{stfloats}
\usepackage{tabularx, makecell, hhline}
\usepackage{fixltx2e}
\usepackage[normalem]{ulem}
\usepackage{placeins}

\usepackage[capitalize]{cleveref}
\crefname{section}{Sec.}{Secs.}
\Crefname{section}{Section}{Sections}
\Crefname{table}{Table}{Tabs.}
\crefname{table}{Table}{Tabs.}
\usepackage{multirow}
\usepackage{tikz}
\usetikzlibrary{positioning,calc,backgrounds}
\usepackage[]{tikzsymbols}
\usepackage[abs]{overpic}
\usepackage{verbatim}
\usepackage{amsmath}
\usepackage{amssymb}
\usepackage{mathtools} 
\usepackage{afterpage}
\newcommand*{\newcite}[1]{~\cite{#1}}

\usepackage{pifont}
\definecolor{ForestGreen}{rgb}{0.13, 0.55, 0.13}

\definecolor{odarkgreen}{HTML}{28B23D}        
\definecolor{odarkred}{HTML}{ED1C1C}            

\usepackage{xcolor}
\usepackage{colortbl}

\usepackage{soul}
\definecolor{lightgray}{rgb}{0.94, 0.94, 0.94}
\definecolor{lgray}{rgb}{0.9, 0.9, 0.9}
\definecolor{lightblue}{rgb}{0.9254901960784314, 0.9529411764705882, 0.9921568627450981} 
\definecolor{lightgreen}{rgb}{0.8980392156862745,1,0.9215686274509803}

\begin{document}

\twocolumn[
\icmltitle{\textit{ETHER}: Efficient Finetuning of Large-Scale Models with Hyperplane Reflections}

\icmlsetsymbol{equal}{*}

\begin{icmlauthorlist}
\icmlauthor{Massimo Bini}{t1,hel}
\icmlauthor{Karsten Roth}{t2,hel}
\icmlauthor{Zeynep Akata}{hel,tum,mcml}
\icmlauthor{Anna Khoreva}{bos}
\end{icmlauthorlist}

\icmlaffiliation{t1}{Bosch IoC Lab, University of Tübingen}
\icmlaffiliation{t2}{Tübingen AI Center, University of Tübingen}
\icmlaffiliation{hel}{Helmholtz Munich}
\icmlaffiliation{tum}{Technical University of Munich}
\icmlaffiliation{bos}{Bosch Center for Artificial Intelligence}
\icmlaffiliation{mcml}{Munich Center for Machine Learning}
\icmlcorrespondingauthor{Massimo Bini}{massimo.bini@uni-tuebingen.de}

\icmlkeywords{Machine Learning, ICML}

\vskip 0.3in
]

\printAffiliationsAndNotice{}  % leave blank if no need to 

\begin{abstract}
Parameter-efficient finetuning (PEFT) has become ubiquitous to adapt foundation models to downstream task requirements while retaining their generalization ability.
However, the amount of additionally introduced parameters and compute for successful adaptation and hyperparameter searches can explode quickly, especially when deployed at scale to serve numerous individual requests.
To ensure effective, parameter-efficient, and hyperparameter-robust adaptation, we propose the \textsl{ETHER} transformation family, which performs \textsl{E}fficient fine\textsl{T}uning via \textsl{H}yp\textsl{E}rplane \textsl{R}eflections. 
By design, \textsl{ETHER} transformations require \textit{a minimal number of parameters}, are \textit{less likely to deteriorate model performance}, and exhibit \textit{robustness to hyperparameter and learning rate choices}.
In particular, we introduce \textsl{ETHER} and its relaxation \textsl{ETHER+}, which match or outperform existing PEFT methods with significantly fewer parameters (\raisebox{0.4ex}{\scalebox{0.6}{$\sim$}}$10$-$100$ times lower than LoRA or OFT) across multiple image synthesis and natural language tasks without \textit{exhaustive hyperparameter tuning}. 
Finally, we investigate the recent emphasis on Hyperspherical Energy retention for adaptation and raise questions on its practical utility. The code is available at \scalebox{0.97}{\url{https://github.com/mwbini/ether}}.
\end{abstract}

    %sec/0_abstract}    
\section{Introduction}
\label{sec:intro}

Recently, large-scale foundation models \citep{bommasani2021foundationmodels} have demonstrated impressive general-purpose capabilities across both generative and discriminative tasks~\newcite{rombach2022SD,Touvron2023LLaMAOA,OpenAI2023GPT4TR,kirillov2023sam}, showing extensive flexibility and strong performance when further adapted to different, more specialized tasks such as instruction following or controlled image synthesis~\newcite{zhang2023controlnet,Ruiz2022DreamBoothFT,alpaca,vicuna2023}.

While impressive, these capabilities come with parameter counts increasing into the billions \newcite{OpenAI2023GPT4TR,Podell2023SDXLIL,Touvron2023Llama2O}. To allow for affordable and scalable model adaptation that can serve large and diverse client bases, various techniques have been introduced in the literature.
They range from full finetuning~\citep{zhao2024galore,zhang2023slca,stojanovski2022momentumbased} to just a few layers of the pretrained model \newcite{kornblith_better_2019}, concatenating additional learning modules \newcite{houlsby_parameter-efficient_2019,pfeiffer_adapterfusion_2020,mou2023t2i-adapter}, and more recently to adapters on the network weights with lightweight learnable transformations \newcite{qiu2023oft,hu2022lora,kopiczko_vera_2023,valipour_dylora_2023}.
The latter have proven particularly effective, introducing no inference latency, fewer adaptation parameters, and strong performance.

Conceptually, these methods finetune on smaller datasets to adapt to downstream task and data requirements, without (1) compromising too much on the costly pretraining and (2) incurring concept and semantic drifts by catastrophically overwriting pretrained weights \citep{kirkpatrick2016catfor,lee2019drift,lu2020drift,mehta2022an,ruiz_dreambooth_2023,ke2023continual,roth2024fantastic,garg2024ticclip,ibrahim2024simple}.
Treading the line for a suitable trade-off between adaptation and retention of the foundational model capabilities thus presents itself as a difficult task to tackle, often requiring costly tuning of hyperparameters such as learning rates.
This problem is acknowledged explicitly in \citet{li_explicit_2018,chen_parameter-efficient_2023,gouk_distance-based_2021} aiming to preserve Euclidean weight distances between pretrained and finetuned models, and implicitly with approaches opting for both lower learning rates (at the cost of more tuning iterations) and inclusion of tuning parameters via summation \citep{qiu2023oft}.

In particular, \citet{qiu2023oft} hints that a Euclidean distance measure likely fails to fully capture the preservation of the network's ability, suggesting instead Hyperspherical Energy (HE) as an alternative measure. The resulting objective uses orthogonal transformations (OFT) for multiplicative weight changes that control HE. Still, even OFT requires specific and restricted hyperparameter choices such as small learning rates and initialization from identity matrices to ensure sufficient knowledge preservation. In addition, while more robust and stable for finetuning in controllable generation settings compared to LoRA \citep{qiu2023oft}, OFT comes with a high computational overhead due to matrix multiplication and a large number of tuning parameters.

In this work, we propose \textbf{E}fficient fine\textbf{T}uning via \textbf{H}yp\textbf{E}rplane \textbf{R}eflections (\textsl{ETHER}) - a new family of weight transformations, efficient in parameter count while preserving model abilities and being robust in convergence and learning rate choices.
By default, \textsl{ETHER} transformations frame the tuning process as a search for suitable hyperplanes, along which weight vectors can be reflected based on the orthogonal Householder transformation \citep{householder1958householder}. This keeps the distance to the transformation neutral element - the identity matrix - constant by construction and improves training stability while reducing the chance of deteriorating model performance. In addition, being built from single vectors, Householder transformations allow for efficient block-parallel matrix multiplication with minimal performance trade-offs.

However, situations may arise where the hard distance restriction of \textsl{ETHER} can prove suboptimal (such as for subject-driven image generation, where finegrained subject-specific semantics need to be retained).
As such, we augment the \textsl{ETHER} family with \textsl{ETHER+} - a relaxation on the default \textsl{ETHER} method. 
More precisely, \textsl{ETHER+} derives from the Householder transformation, but breaks the orthogonality and constant distance constraints, introducing multiple hyperplanes that can interact with a weight vector.
As a result, \textsl{ETHER+} allows for more controlled and finegrained adaptation, while still having a bounded distance to the transformation neutral element, and retaining the \textsl{ETHER} benefits of high parameter-efficiency, training stability, and hyperparameter robustness.

Indeed, across subject-driven image generation, controlled image synthesis, natural language understanding and instruction tuning tasks, we find that \textsl{ETHER} and especially \textsl{ETHER+} match and outperform existing methods using only a few additional tuning parameters (e.g. $100\times$ less than OFT when finetuning Stable Diffusion for controlled image synthesis) - all while presenting stronger learning rate robustness compared to other methods and consequently requiring minimal hyperparameter tuning to achieve strong performance (c.f. Sec.~\ref{sec:intrigue}).
Finally, we also utilize our experimental benchmark findings to further investigate and question the recent emphasis on transformation orthogonality and hyperspherical energy (HE) retention (e.g. \citet{qiu2023oft}), showing how non-orthogonal \textsl{ETHER+} can achieve strong performance while displaying increased HE.
       %sec/1_intro}
\section{Related Work}
\label{sec:relatedwork}

\textbf{Parameter-Efficient Finetuning (PEFT).} PEFT of pretrained models has seen different strategies evolve in the past years - starting from finetuning protocols and concatenation of learnable modules~\newcite{houlsby_parameter-efficient_2019,lester_power_2021-1,li2021prefixtuning,pfeiffer_adapterfusion_2020,guo2021parameterefficient} to more recently reparametrization of network weights with efficient transformations~\newcite{qiu2023oft,hu2022lora,kopiczko_vera_2023,valipour_dylora_2023,zhang_adaptive_2023}.
The latter have shown convincing trade-offs between adaptation quality, additional parameters, and inference latency. LoRA \newcite{hu2022lora} transforms network weights by adding the result of a learnable, low-rank matrix product. On top of LoRA, multiple variations have been proposed, s.a. QLora~\citep{dettmers_qlora_2023} with quantized weights, AdaLoRA~\citep{zhang_adaptive_2023} with dynamic rank adjustment, and VeRA \cite{kopiczko_vera_2023} with low-rank frozen random projections and trainable vectors to reduce parameter counts. 
OFT \newcite{qiu2023oft} instead learns matrix multiplier with orthogonality constraints to retain hyperspherical energy. 
In our work, we use the same paradigm but introduce hyperplane reflections for better parameter efficiency and learning rate robustness.

\textbf{Controlling Diffusion Generative Models.} Diffusion-based generative models show strong compositional generation \citep{rombach2022SD,mukhopadhyay_diffusion_2023,podell_sdxl_2023,karthik2023dont,saharia_photorealistic_2022}. Among these, \citet{gal2022image, ruiz_dreambooth_2023} popularized personalized generation - teaching models to generate variations of user-provided samples. Based on DreamBooth \citep{ruiz_dreambooth_2023}, other works \newcite{liu_cones_2023,richardson_conceptlab_2023,zhang_inversion-based_2023} followed. ControlNet \citep{zhang_adding_2023} shows model controllability through external signals s.a. semantic and depth maps or face landmarks via extra layers at the cost of higher inference latency.
\citet{qiu2023oft} show controllability through direct finetuning with learnable matrix-multiplication transformations.
Our work suggests an alternative, more robust and parameter-efficient approach through hyperplane reflections.

\textbf{Instruction Tuning Language Models.} Large Language Models (LLMs) have shown striking generalization across a wide range of tasks \citep{zhao_survey_2023,zhang_instruction_2023,OpenAI2023GPT4TR,Touvron2023LLaMAOA}. However, the default training objective often does not exactly match downstream task requirements and intentions.
To address this mismatch, Instruction Tuning \citep{wang_self-instruct_2023,zhang_instruction_2023,longpre_flan_nodate,alpaca} finetunes LLMs using additional \texttt{(Instruction, Output)} pairs to explicitly align the model with human preferences.
This enhances capabilities and controllability while avoiding costly retraining \citep{köpf2023openassistant}. Recently, methods based on LoRA \cite{hu2022lora} have been proposed to efficiently achieve this control \cite{dettmers_qlora_2023,xu_qa-lora_2023,chen_longlora_2023,valipour_dylora_2023,kopiczko_vera_2023}.
This work proposes a strong alternative with further parameter-efficiency and high learning rate robustness.

  %sec/2_related_works}
\section{Method}
\label{sec:3_method}

We first discuss adapter-based PEFT in \S\ref{subsec:prelims}, before describing and motivating the use of hyperplane reflections in \textsl{ETHER} (\S\ref{subsec:ether}). To encourage flexibility in trainable control and adaptation, we propose a simple, yet effective relaxation \textsl{ETHER+} in \S\ref{subsec:ether+}. Finally, \S\ref{subsec:blockparallel} describes block-diagonal \textsl{ETHER} for improved computational efficiency.

\subsection{Preliminaries} \label{subsec:prelims}
\paragraph{Parameter-Efficient Finetuning with Adapters.} The most commonly deployed form of PEFT with an adapter is \textit{Low-rank Adaptation} (\textit{LoRA}, \citet{hu2022lora}). LoRA parametrizes a change of pretrained weights $W$ as 
\begin{equation*}
(W+BA)^{\intercal}x + b \\[1pt]  
\end{equation*}
where $BA$ is the matrix product of two low-rank matrices, i.e. for $W\in\mathbb{R}^{d\times f}$, $A\in\mathbb{R}^{d\times r}$ and $B\in\mathbb{R}^{r\times f}$. When rank $r << \text{min}(d,f)$, this can bring down required tuning parameters significantly compared to full finetuning. In addition, $BA$ can be absorbed into $W$ during inference to avoid additional latency.

\paragraph{Orthogonal Finetuning (OFT).}
However, finetuning with LoRA can incur significant, potentially catastrophic weight changes. To ensure better preservation of pretrained model weights, \citet{qiu2023oft} propose Orthogonal Finetuning (OFT). Based on the hypothesis that Hyperspherical Energy (HE) needs to be kept unaltered to preserve the original model abilities, OFT proposes the usage of multiplicative orthogonal transformations on the model weights. By retaining pairwise weight angles, HE can remain unaffected. 
However, to work in practice, \citet{qiu2023oft} require the construction of the orthogonal matrix $Q$ via a Cayley parametrization $Q=(I+S)(I-S)^{-1}$, where $S$ is skew-symmetric. Notice that by using this parametrization, they limit the range of possible orthogonal matrices to those with determinant 1, missing orthogonal matrices with determinant equal to $-1$. As we show, this is relevant, as it excludes reflections, which motivate \textsl{ETHER}.
To make OFT more parameter efficient, the orthogonal matrix $Q\in\mathbb{R}^{d\times d}$ is built in a block-diagonal fashion, made up of $n$ smaller blocks $Q^b$ of size $\frac{d}{n}\times \frac{d}{n}$.
The final OFT transformation on the forward pass can then be described as
\begin{equation*}
(Q^BW)^{\intercal}x+b    
\end{equation*}
with block-diagonal $Q^B$. The trainable parameters are the $n$ matrices $Q^b\in\mathbb{R}^{\frac{d}{n} \times \frac{d}{n}}$ that compose $Q^B$ - more specifically the matrices $R^b$ that build the skew-symmetric matrices $S^b=\frac{1}{2}(R^b-(R^b)^{\intercal})$ for $Q^b$.
For finetuning, the $R^b$ are initialized as zero, such that $Q^B|_0=I$ and consequently $Q^B|_0W = W$ at the beginning of finetuning.

\subsection{\textsl{ETHER}: Finetuning with Hyperplane Reflections}\label{subsec:ether}

\begin{figure}[t]
  \centering
  \vspace{0.3cm}
   \includegraphics[width=1\linewidth]{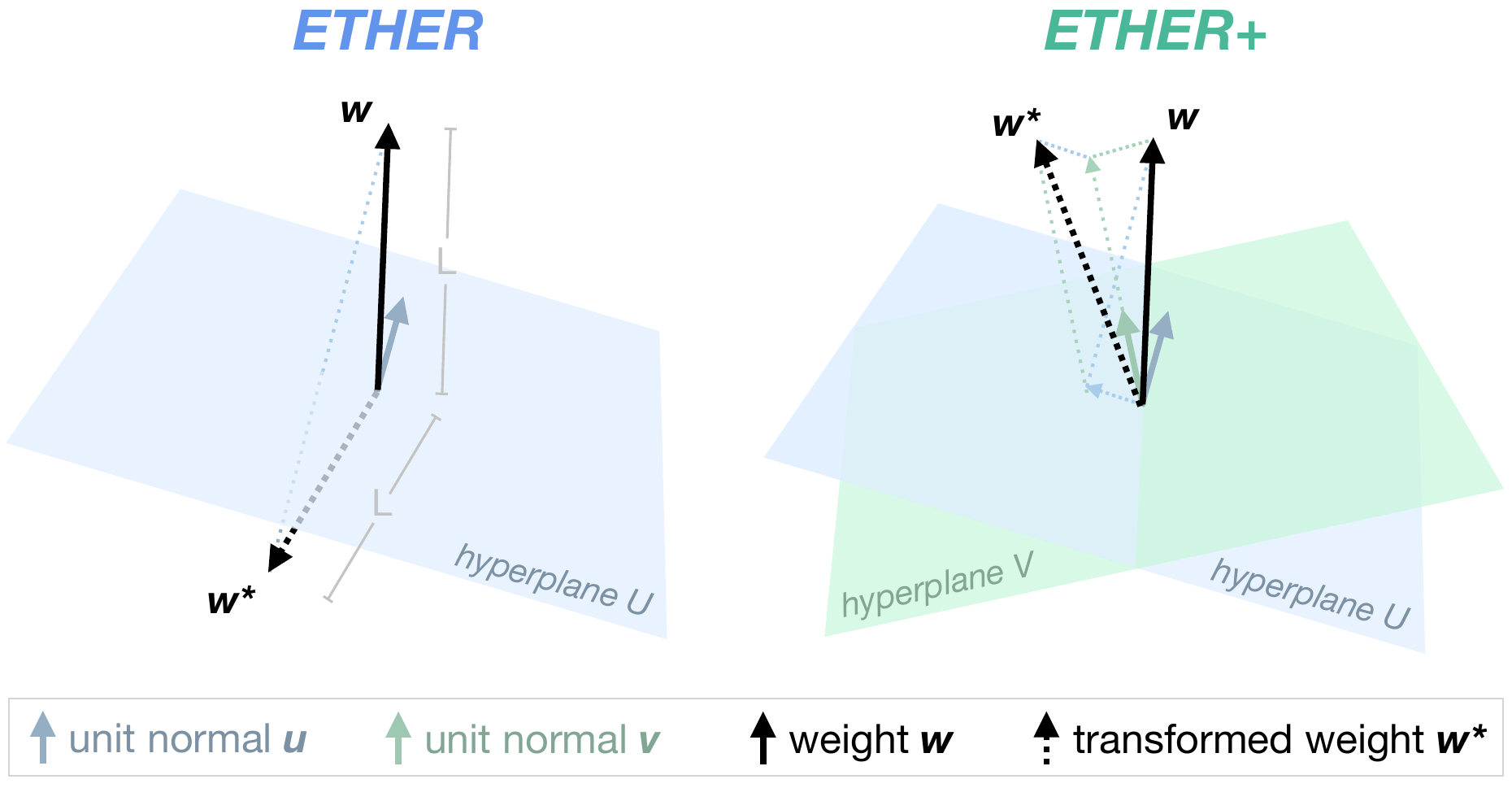}
   \vspace{-0.3cm}
   \caption{\textbf{\textsl{ETHER} and \textsl{ETHER+} sketches.} We visualize either a single hyperplane reflection for \textsl{ETHER} or two interacting hyperplanes for \textsl{ETHER+}, parametrized unit normals $u$ (and $v$). Unlike \textsl{ETHER}, the final result of \textsl{ETHER+} does not have to retain the original length $L$, as the need for hard reflections is softened, and orthogonality is no longer guaranteed.}
   \label{fig:ether_sketch}
   \vspace{-0.2cm}
   % \vspace{-5pt}
\end{figure}

Fundamentally, \textit{ETHER} (\textbf{E}fficient fine\textbf{T}uning via \textbf{H}yp\textbf{E}rplane \textbf{R}eflections) sets up weight transformations as hyperplane reflections. These reflections can be obtained via the Householder transformation matrix $H\in\mathbb{R}^{d\times d}$ with
\begin{equation}\label{eq:householder}
    H=I-2uu^{\intercal}
\end{equation}
with $u\in\mathbb{R}^d$ the hyperplane unit normal vector and the corresponding outer product $uu^{\intercal}$.
The reflection can be easily intuited when applied to a weight vector $w\in\mathbb{R}^d$:
\begin{equation*}
Hw = (I - 2uu^{\intercal})w = w - 2u(u^{\intercal}w).
\end{equation*}
Transformation $H$ effectively subtracts twice the component of $w$ projected on $u$, thereby reflecting it with respect to the hyperplane defined by $u$ (see Fig.~\ref{fig:ether_sketch}).
By construction, hyperplane reflections are well-suited for the efficient finetuning of pretrained models, as they keep the distance to the transformation neutral element - the identity matrix - constant, which minimizes the risk of divergence from the pretrained model and deterioration of model performance (c.f. Fig.~\ref{fig:distance_vs_lr}). 
This can be easily shown by computing the Frobenius norm of the difference between the Householder matrix H and the identity matrix I:
\begin{equation}\label{eq:frob}
\left\Vert H - I\right\Vert_F = \left\Vert I-2uu^{\intercal}-I \right\Vert_F = 2\cdot \left\Vert uu^{\intercal} \right\Vert_F = 2
\end{equation}
The above equation leverages the fact that for any matrix $M$
\begin{equation*}
    \left\Vert M \right\Vert_F = \sqrt{\text{Tr}(MM^\intercal)}    
\end{equation*}
and that with $M = uu^\intercal$ and $u$ having unit length $u_1^2 + u_2^2 + ... + u_d^2 = 1$, one can simply write (with $(uu^\intercal)^\intercal = uu^\intercal$)
\begin{equation*}
\textstyle \left\Vert uu^\intercal\right\Vert_F = \sqrt{\sum_{i=1}^d u^2_i} = 1.
\end{equation*}

Since the finetuning process simply consists of finding the optimal directions of the reflection hyperplanes with bounded deviations from the transformation neutral element, it allows for (i) a very \textit{low number of extra parameters} corresponding to the unit vectors $u$, and (ii) the usage of high learning rates, as \textit{the risk of divergence is minimized}. This allows for general \textit{learning rate robustness} and encourages fast convergence by default, as consistently high learning rates can be selected; reducing computational resources required to achieve good performance (e.g. Fig.~\ref{fig:convergence_cn}).

Interestingly, as this transformation is orthogonal ($HH^\intercal = I$), it falls under the umbrella of orthogonal transformations motivated in OFT~\citep{qiu2023oft} from the perspective of Hyperspherical Energy control to better preserve model pretraining. However, OFT leverages the Cayley parametrization of orthogonal matrices, which only produces determinant 1 matrices. By construction, this excludes Householder matrices from OFT, which have determinant $-1$!
However, as noted above, it is indeed in this particular setting and through the use of Householder transformations that high parameter efficiency, strong pretraining retention, and learning rate robustness arise.

On top of that, we further investigate the importance of Hyperspherical Energy retention by conducting a control study comparing OFT against its non-orthogonal variant (\textit{Naive})\footnote{\textit{Naive} employs an unconstrained block-diagonal transformation matrix $N^{B}$ made up of $n$ blocks and initialized as an identity matrix, i.e. having the same number of trainable parameters and initialization as OFT's  transformation matrix $Q^B$.}
Our experiments do not show significant differences in terms of control and training stability, suggesting that such properties stem from the multiplicative finetuning approach rather than the underlying HE retention, contrasting insights in \citet{qiu2023oft} (c.f. Sec.~\ref{subsec:orthogonality}).
These findings partly motivate the exploration of a relaxed variant of the Householder reflection in the next section~\ref{subsec:ether+}, which demonstrates that loosening the orthogonality constraint not only maintains good performance but can even lead to enhanced results.

\subsection{Relaxing Orthogonality in \textsl{ETHER}}\label{subsec:ether+}
\begin{figure}[t]
  \centering
  \vspace{0.1cm}
   \includegraphics[width=\linewidth]{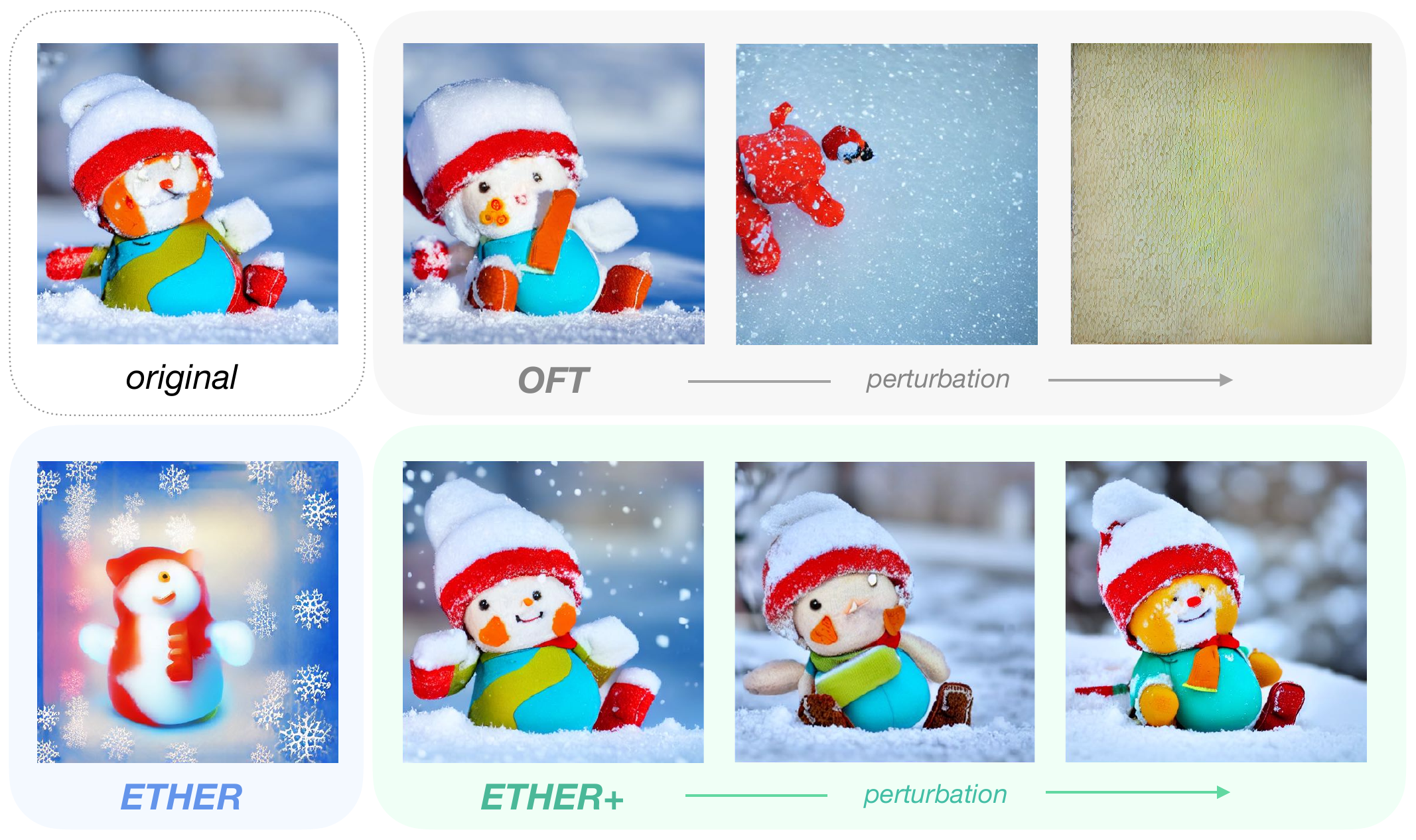}
   \vspace{-0.3cm}
   \caption{\textbf{Change in model behavior as a function of perturbation strength}, i.e. distance between weight transformation and identity matrix. As \textsl{ETHER} and \textsl{ETHER+} are upper-bounded in perturbation by construction, catastrophic deterioration of model performances is rarely encountered, and weight transformations remain controllable even for maximal deviations. For standard approaches, s.a. OFT, larger deviations from the identity matrix may occur during training and result in substantial divergence from the pretrained model. Notice also that by breaking orthogonality constraints in \textsl{ETHER+}, both smaller and stronger semantic variants can be learned.}
   \label{fig:sample_perturbations}
   \vspace{0.1cm}
\end{figure}

\begin{figure}[t]
  \centering
  \vspace{0.2cm}
   \includegraphics[width=1\linewidth]{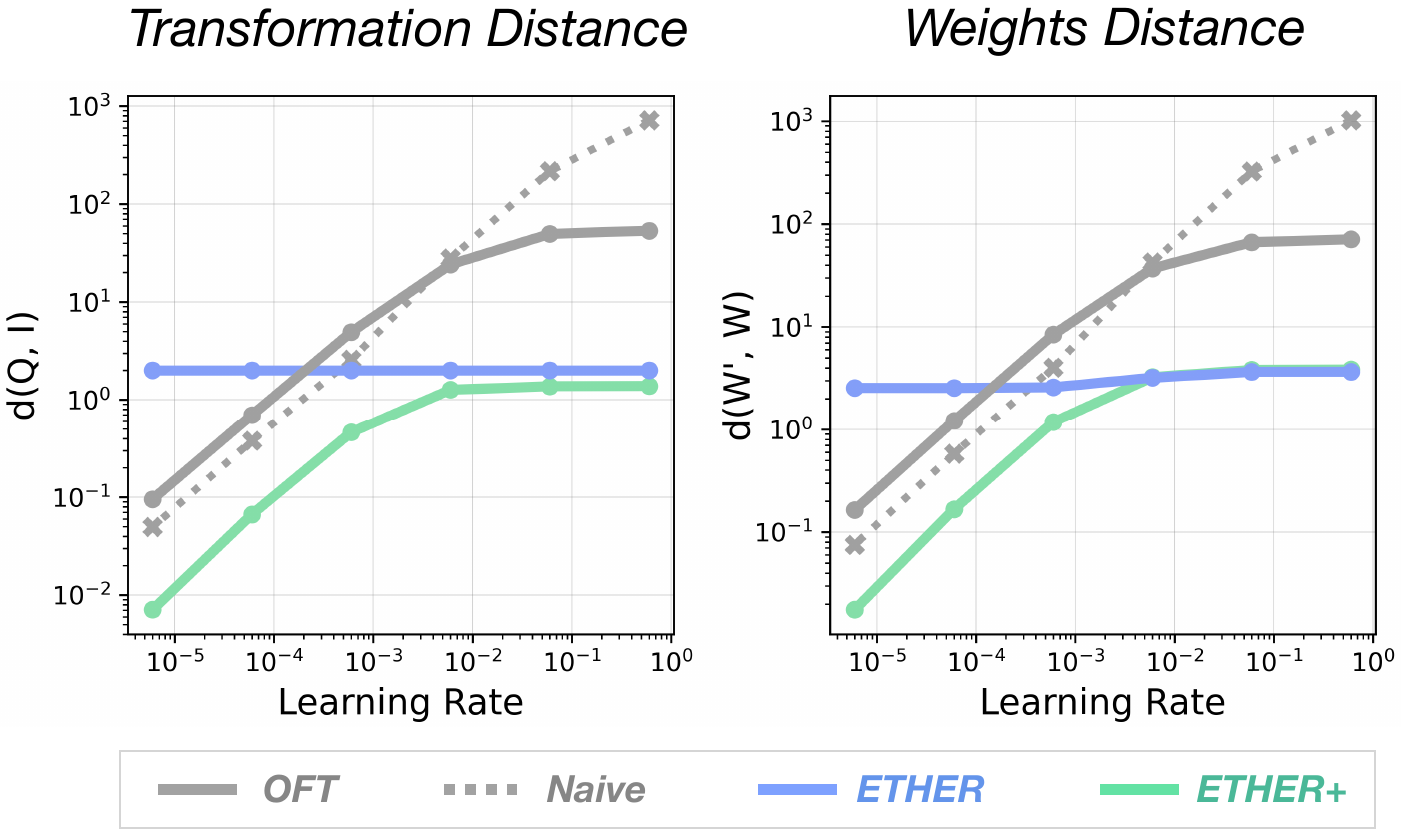}
   \vspace{-0.3cm}
   \caption{\textbf{Distances as a function of learning rates} between transformation and identity matrix (\textit{Transformation Distance}), and finetuned and pretrained weights (\textit{Weights Distance}). Distances obtained for subject-driven generation finetuning at convergence (1200 iterations). Results show distances magnitudes higher and unbounded for non-\textsl{ETHER} methods in both cases as learning rates increase.}
   \label{fig:distance_vs_lr}
   \vspace{-0.2cm}
\end{figure}

While finetuning via hyperplane reflections has several promising qualities as highlighted above, there is no free lunch. In particular, situations may arise where the strength of the transformation and inherent deviation from the identity may be too large by default, such as for potentially more nuanced tasks like subject-driven generation \citep{ruiz_dreambooth_2023}. To allow for more nuanced transformations while retaining beneficial properties of \textsl{ETHER} - parameter efficiency and learning rate robustness through bounded deviations from the transformation neutral element - we propose the \textsl{ETHER+} relaxation
\begin{equation*}\label{eq:dir}
    H^+=I-uu^{\intercal}+vv^{\intercal}
\end{equation*}
with unit vectors $u, v\in\mathbb{R}^d$. This is a simple variation of the Householder transformation that now allows for interaction between two distinct hyperplanes (see Fig.~\ref{fig:ether_sketch}). This helps to control the transformation strength as $uu^\intercal$ and $vv^\intercal$ can weaken or even cancel each other out to return the identity transformation in the limit where $u = v$.
In addition, the transformation distance remains bounded, as the relaxed variant $H^+$ always has $\left\Vert H^+ - I \right\Vert_F \leq 2$, i.e. 
\begin{equation*}\label{eq:relaxed}
\text{max} \left\Vert H^+ - I \right\Vert_F \leq \text{max}\left\Vert H - I \right\Vert_F. 
\end{equation*}
This follows immediately from the triangle inequality of norms, i.e. $\left\Vert vv^\intercal - uu^\intercal \right\Vert_F \leq \left\Vert vv^\intercal \right\Vert_F + \left\Vert uu^\intercal\right\Vert_F = 2$.
Due to the weaker strength of this new transformation, we apply it both on the left ($H^+$) and right ($\tilde{H}^+$) of the weight matrix $W$, such that the forward pass becomes 
\begin{equation*}
\left(H^+ W \tilde{H}^+\right) ^\intercal x+b.
\end{equation*}
Consequently, \textsl{ETHER+} effectively leverages a sequence of hyperplane interactions that no longer have to retain length to allow for more nuanced weight adjustment while still minimizing the risk of diverging from the pretrained model (as also shown e.g. in Figs.~\ref{fig:sample_perturbations}, ~\ref{fig:distance_vs_lr},~\ref{fig:miou_vs_lr} and~\ref{fig:convergence_cn}).

\begin{figure}[t]
  \centering
  \vspace{0.2cm}
   \includegraphics[width=1\linewidth]{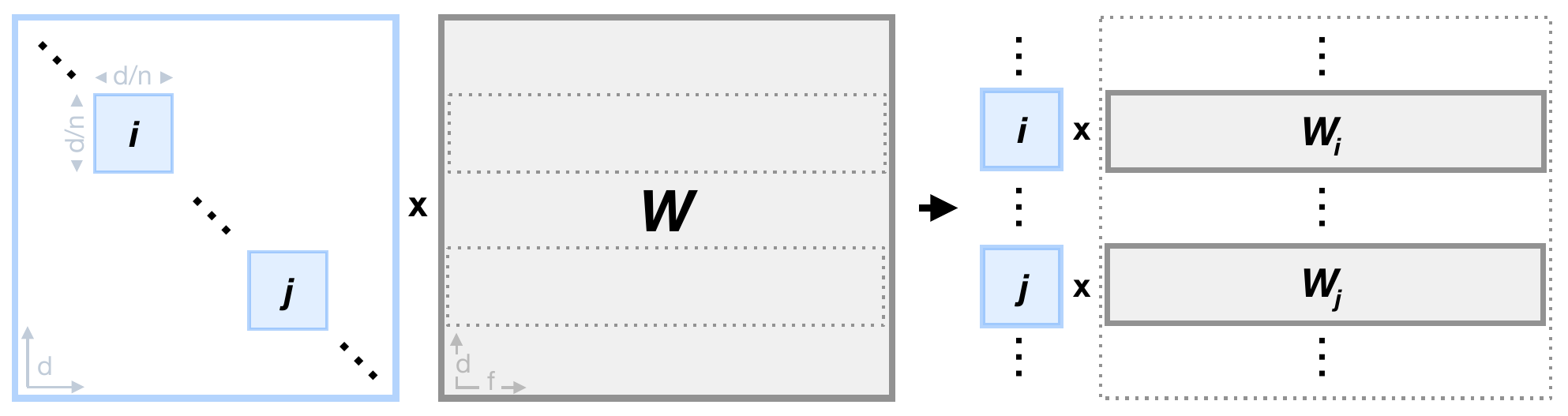}
   \vspace{-0.3cm}
   \caption{\textbf{Block-Parallel Computation scheme} between $d$-dimensional block-diagonal transformation with $n$ blocks and a $d\times f$ -dimensional weight matrix $W$.}
   \label{fig:parallel_both}
   \vspace{-0.05cm}
\end{figure}

\begin{table}[t]
\centering
\vspace{-0.1cm}
\caption{\textbf{Better computational efficiency through block-diagonality} on \textbf{Phi1.5}-1.3B and \textbf{Llama-2}-7B, with internal dimensions of 2048 and 4096 respectively. As the number of blocks $n$ increases, so does the computational efficiency, quantified by the decrease in TFLOPs required for a single backward pass (using a sample with longest sequence length). The larger the model's internal dimension, the larger the efficiency gain.}
\vspace{0.3cm}
\resizebox{1\linewidth}{!}{
\begin{tabular}{l|cc|cc}
\multirow{2}{*}{} & \multicolumn{2}{c|}{\textbf{Phi1.5-1.3B}} & \multicolumn{2}{c}{\textbf{Llama-2-7B}} \\ 
& TFLOPs      & rel. drop  & TFLOPs & rel. drop \\ 
 \toprule
\rowcolor{lightgray}LoRA$_{r=8}$ & 6.04 & - & 6.85& -\\
\rowcolor{lightgray}OFT$_{n=256}$ & 9.13 & - & 25.26& -\\
 \textsl{ETHER}$_{n=1}$ & 9.13 & - & 25.26& -\\
\rowcolor{lightblue} \textsl{ETHER}$_{n=4}$ &  7.07 & -23\%  & 12.07 & -52\% \\
\rowcolor{lightblue} \textsl{ETHER}$_{n=32}$ & 6.71 & -27\% & 8.22 & -68\% \\
 \textsl{ETHER+}$_{n=1}$ & 10.78 & - & 51.65& -\\
\rowcolor{lightgreen} \textsl{ETHER+}$_{n=4}$ &  7.69 & -29\%  & 18.66 & -64\% \\
\rowcolor{lightgreen} \textsl{ETHER+}$_{n=32}$ & 6.79 & -37\% & 9.04 & -83\% \\
\end{tabular}
}
\label{tab:load}
\end{table}

\subsection{Efficient \textsl{ETHER} through Block-Parallelism}\label{subsec:blockparallel}

In multiplicative finetuning like OFT or \textsl{ETHER}, further computational load is introduced through additional matrix multiplications. 
To mitigate this issue, we introduce a block-diagonal formulation of \textsl{ETHER} similar to block-diagonal OFT described in \S\ref{subsec:prelims}. 
For this, we break down the Householder transformation $H$ (eq.~\ref{eq:householder}) into its corresponding block-diagonal variant $H^B$:
\begin{equation*}\label{eq:blockdiag}
    \text{diag}(H^1 \cdots H^n) = I - 2
\begin{pmatrix}
\hat{u}_1\hat{u}_1^\intercal & & \\
& \ddots  &  \\
& & \hat{u}_n\hat{u}_n^\intercal\\
\end{pmatrix}
\end{equation*}
with each $i$-th block-plane parameterized by $\hat{u}_i\in\mathbb{R}^{\frac{d}{n}}$. Of course, one can do the same for $H^+$. In both cases, such a block-diagonal formulation reduces the cost of computing $H$.
More importantly, each $i$-th block now only affects the corresponding $i$-th block-row in the weight matrix $W$. This means we can split $W$ into $n$ sub-blocks $W^i\in\mathbb{R}^{\frac{d}{n}\times f}$, each of which is uniquely altered by its corresponding $H^i$ counterpart. As a result, the full weight transformation can now be separated into smaller block-specific operations, reducing the overall number of computations. Furthermore, these operations can now be fully block-parallelized, significantly increasing training speed!
In terms of computations, for each full-matrix-multiplication between $H$ and $W$ of sizes $d\hspace{-0.05cm}\times\hspace{-0.05cm}d$ and $d\hspace{-0.05cm}\times\hspace{-0.05cm}f$ respectively, $d(df)$ multiplications and $(d\hspace{-0.05cm}-\hspace{-0.05cm}1)df$ additions are necessary, accounting for $\mathcal{O}(d^2 f)$ operations. With our block-parallel scheme, we reduce these to $n$ block-specific $\frac{d}{n}(\frac{d}{n}f)$ multiplications and $\frac{d-1}{n}(\frac{d}{n}f)$ additions, resulting in $\mathcal{O}(\frac{d^2 f}{n})$ operations (see Tab. \ref{tab:load}).

Furthermore,  with each block being built from a single vector of dimension $\frac{d}{n}$, \textsl{ETHER} transformations' construction ensures that the total number of trainable parameters remains constant for any $n$ number of blocks.
This stands in contrast to block-diagonal OFT, where the use of higher block counts was introduced to minimize the number of parameters while introducing noticeable decreases in adaptation performance!
Instead, for block-diagonal \textsl{ETHER}, we find performance to be consistent over increasing block counts (see App.~\ref{suppsec:ablations}), allowing for an improved computational footprint with negligible performance decrease.

%%%%%%%%%%%%%%%%%%%%%%%%%%%%%%%%%%%%

\section{Intriguing Properties of ETHER}\label{sec:intrigue}

This section investigates and highlights the bounded distance and non-deteriorating nature of \textsl{ETHER}/\textsl{ETHER+} in more detail while providing insights into its favorable learning rate robustness and the reliable use of high learning rates for fast convergence. For completeness, we also report here comparisons with the unconstrained \textit{Naive} method, to better show the impact of orthogonality as proposed by \citet{qiu2023oft}, and how our method provides much stronger robustness. Finally, we include a discussion on the parameter efficiency. For all experiments in this section, please see \S\ref{subsec:generative_models_adaptation} for relevant implementation details.

\textbf{Non-Deteriorating Nature.} Because both \textsl{ETHER} and \textsl{ETHER+} are upper-bounded in their possible perturbation over the pretrained weight matrices (as measured for example by the distance to the transformation neutral element, the identity matrix), finetuning with both methods will guarantee suitable results for most hyperparameter choices.
This is easily visualized in Fig.~\ref{fig:sample_perturbations} by looking at generation samples after perturbing Stable Diffusion with randomly sampled transformations for each approach - OFT, \textsl{ETHER} and \textsl{ETHER+} - respectively.
While \textsl{ETHER} uses a fixed-distance transformation (c.f. Eq.~\ref{eq:frob}) that introduces a noticeable change (but still retaining semantics), \textsl{ETHER+} can obtain both finegrained visual control as well as stronger semantic changes. Conversely, unbounded methods like OFT catastrophically deteriorate a model's generative abilities as the perturbation strength increases.

This results in a much more controlled generation setting for \textsl{ETHER} and \textsl{ETHER+} finetuning. This is also depicted quantitatively in Fig.~\ref{fig:distance_vs_lr}, which shows distances between the learned transformation and the transformed weights (at convergence) to the identity matrix and the pretrained weights, respectively, as a function of the learning rate. As can be seen, larger learning rate values for OFT and \textit{Naive} finetuning (OFT without orthogonality constraints) result in distances that are orders of magnitude higher than those of \textsl{ETHER} and \textsl{ETHER+}, leading to catastrophic deterioration and model collapse (see Fig.~\ref{suppfig:rate_db} in App.).

\begin{figure}[t]
  \centering
  \vspace{0.15cm}
   \includegraphics[width=1\linewidth]{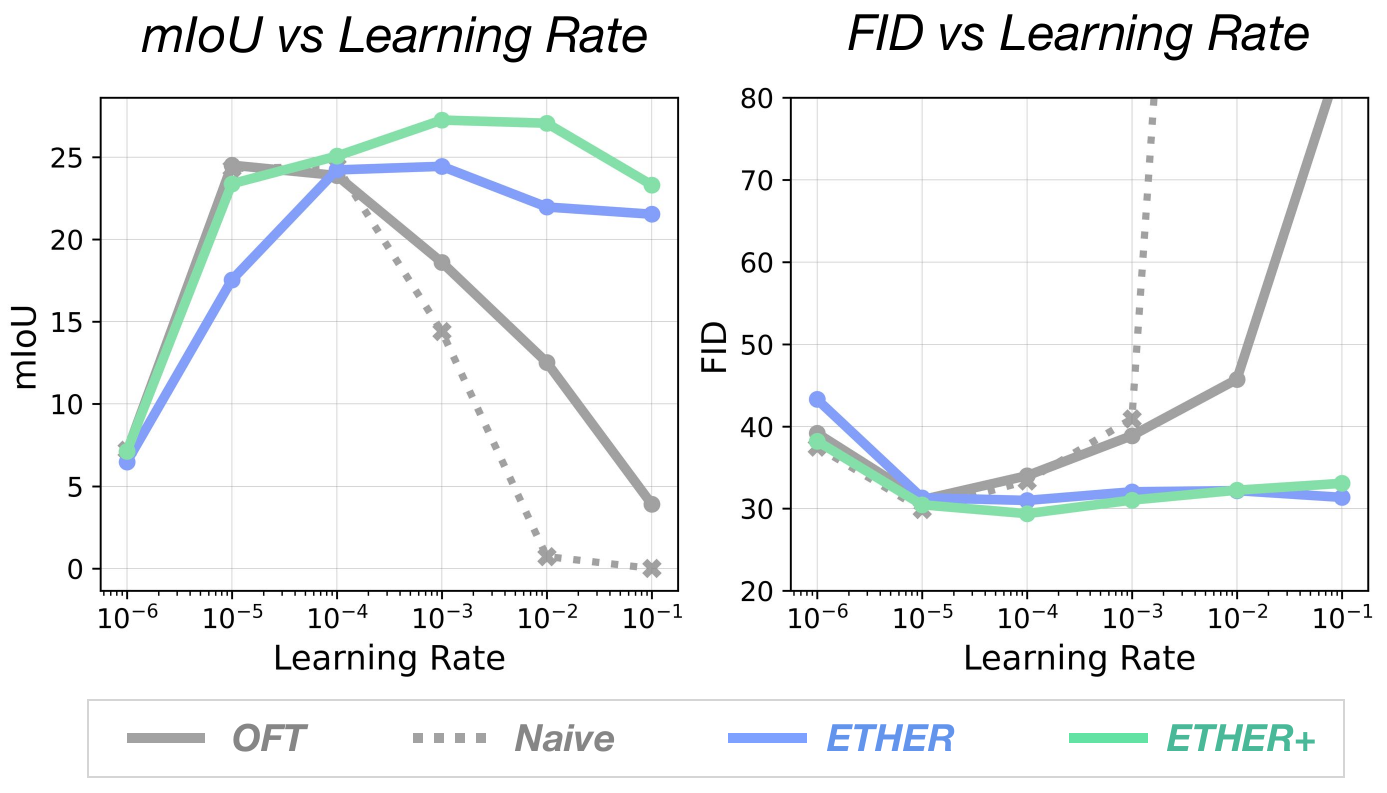}
   \vspace{-0.4cm}
   \caption{\textbf{mIoU and FID performances as a function of learning rates.} Results are obtained for controllable generation S2I finetuning on Stable Diffusion, and reveal a much stronger learning rate robustness of \textsl{ETHER}-based methods; retaining strong performance across entire learning rate magnitudes.}
   \label{fig:miou_vs_lr}
   \vspace{-0.0cm}
\end{figure}

\begin{figure}[t!]
  \centering
  \vspace{0.05cm}
   \includegraphics[width=1\linewidth]{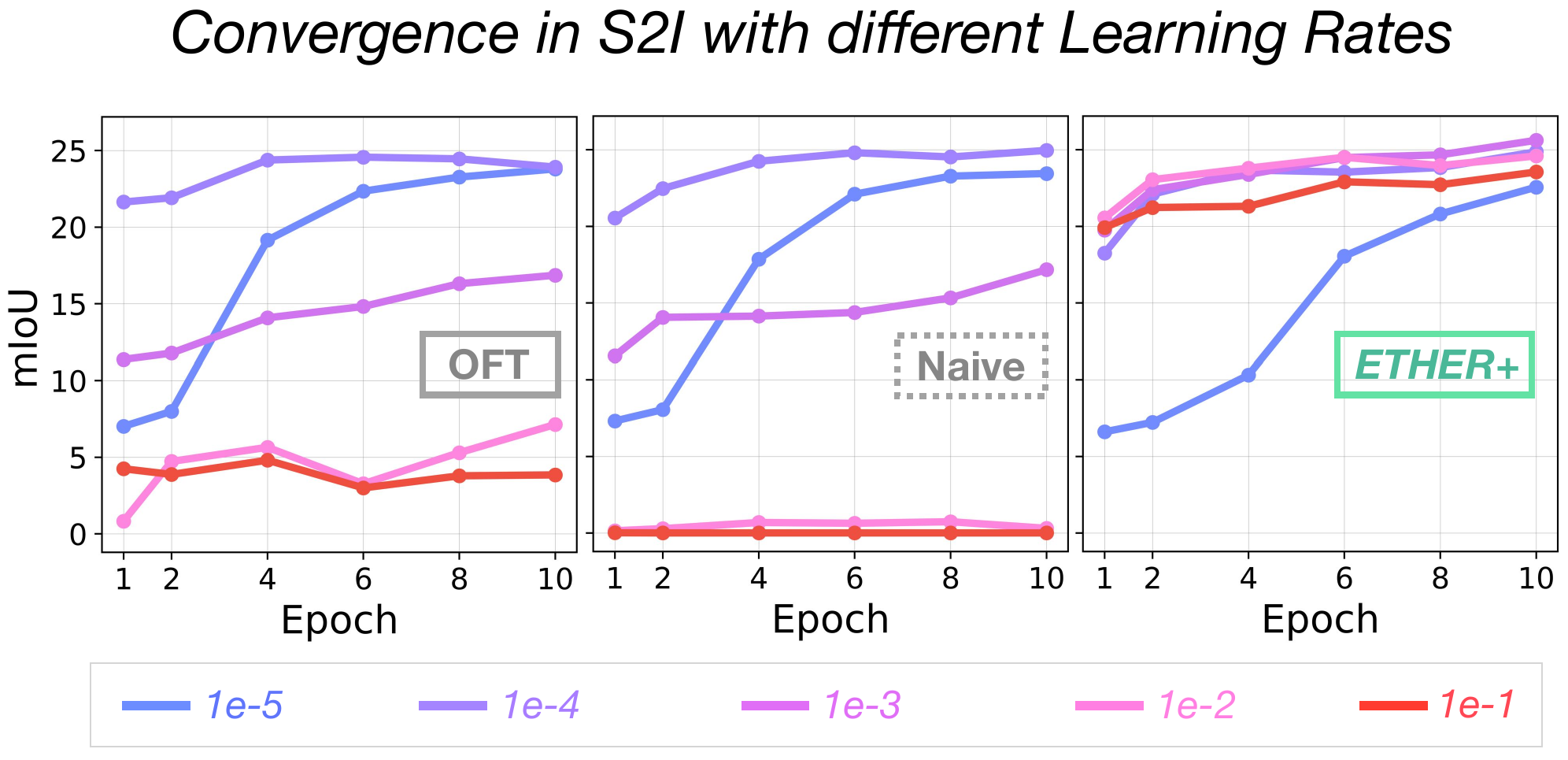}
   \vspace{-0.4cm}
   \caption{\textbf{Achieved controllability (mIoU) per epoch for different finetuning methods.} This figure extends Fig.~\ref{fig:miou_vs_lr} and highlights in detail how only a learning rate of $10^{-4}$ allows for optimal convergence in OFT and Naive, while for \textsl{ETHER+} fastest convergence speeds are stably achieved across magnitudes.}
   \label{fig:convergence_cn}
   \vspace{-0.1cm}
\end{figure}

\textbf{Learning Rate and Hyperparameter Robustness.} Practically, the non-deteriorating nature of \textsl{ETHER} and \textsl{ETHER+} manifests in learning rate robustness during finetuning. As the risks of divergence and collapse are minimized, training stability becomes much less dependent on the choice of learning rate. This is seen when evaluating performance (e.g. mIoU for controllable image synthesis in Fig.~\ref{fig:miou_vs_lr}) and model convergence (Fig.~\ref{fig:convergence_cn}) against learning rates. For non-\textsl{ETHER} methods, Fig.~\ref{fig:miou_vs_lr} shows significant performance drops for high learning rates, while Fig.~\ref{fig:convergence_cn} reveals fast convergence speeds for \textsl{ETHER+} with learning rates covering multiple magnitudes, much more general than e.g. OFT. 

This means that not only can good performance be guaranteed for most learning rate choices, but fast convergence as well, with competitive results already after the first epoch. 
Since \textsl{ETHER} also only introduces a single hyperparameter, the number of diagonal blocks, which marginally impacts performance (c.f. \S\ref{subsec:blockparallel}), \textsl{ETHER} methods become very attractive for practical usage, as the need for grid-search and cautious low learning rate training for good performance (c.f. \S\ref{sec:intro}) is reduced.

\textbf{Parameter Efficiency.} Finally, we provide a more detailed exploration on the parameter efficiency of \textsl{ETHER}-based methods. Let $L$ be the number of finetuned layers, $d$ and $f$ the respective weight dimensions for $W\in\mathbb{R}^{d\times f}$. Then the parameter complexity for OFT can be written as $\mathcal{O}(\frac{Ld^2}{n})$ \citep{qiu2023oft} with $n$ number of diagonal blocks\footnote{\citet{qiu2023oft} note a possible $\mathcal{O}(Ld)$ if $n = \alpha d$. However, in practice, equally scaling $n$ with $d$ disproportionally reduces adaptation parameters for large weight matrices. As OFT is fairly dependent on the parameter count, we omit this estimate.}. Similarly, for LoRA we get $\mathcal{O}(Lr(d+f))$, while for \textsl{ETHER} and \textsl{ETHER+} we only have $\mathcal{O}(Ld)$ and $\mathcal{O}(L(d+f))$ respectively. With respect to both LoRA and OFT, this omits at the very least the rank multiplier $r$, or a potentially quadratic scaling.
As already motivated in Sec.~\ref{sec:3_method}, this results in incredibly efficient finetuning while achieving comparable or stronger performances. For example, when finetuning Stable Diffusion as done above, \textsl{ETHER} and \textsl{ETHER+} use 120 times and 30 times fewer parameters than OFT respectively.

   %sec/3_method}
\section{Benchmark Experiments}
\label{sec:4_experiments}

We first investigate generative model adaptation in Sec.~\ref{subsec:generative_models_adaptation}, with a focus on subject-driven image synthesis (\S\ref{subsubsec:subjectgen}) and controllable image synthesis (\S\ref{subsubsec:controllablegen}) following recent works \citep{qiu2023oft,liu2023parameterefficient}. Sec.~\ref{subsec:language_models_adaptation} then correspondingly investigates language model adaptation, looking at both natural language understanding (\S\ref{subsubsec:langunderstand}) and instruction tuning (\S\ref{subsubsec:instruction_tuning}). Finally, we study the importance of orthogonality and hyperspherical energy on finetuning performance in Sec.~\ref{subsec:orthogonality}.

\subsection{\textsl{ETHER} for Image-generative Model Adaptation}
\label{subsec:generative_models_adaptation}

For our experiments on diffusion-based generative models, we apply the finetuning methods on the pretrained Stable Diffusion-v1.5~\citep{rombach2022SD}, following the setting from OFT \citep{qiu2023oft}. Our experiments follow best practices and hyperparameter choices for each method. For implementation details, please refer to App.~\ref{supp:exp_details}.

\subsubsection{Subject-driven Generation}\label{subsubsec:subjectgen}
We first deploy \textsl{ETHER} and \textsl{ETHER+} on subject-driven generation following \citet{ruiz_dreambooth_2023,qiu2023oft}; finetuning the generative model for each of the 30 subjects and 25 prompts.
For each combination, we generate four images, and measure image quality via a DINO \newcite{caron_emerging_2021} and a CLIP image encoder \newcite{radford_learning_2021}, text-prompt fidelity via a CLIP text encoder, and image diversity using LPIPS \newcite{zhang_unreasonable_2018}. 

\textit{Quantitative Results.} Results are shown in Tab.~\ref{tab:dreambooth}. On subject-driven generation, we find competitive performance for both image quality, text-prompt fidelity and image diversity, particularly for \textsl{ETHER+} (e.g. DINO and CLIP-I scores of $0.666$ vs $0.652$ and $0.8$ vs $0.794$ for OFT, respectively). Most importantly, we achieve this performance while only utilizing a fraction of tuning parameters; with \textsl{ETHER+} only introducing $0.4M$ as compared to $11.6M$ by OFT.
As hypothesized in Sec.~\ref{sec:3_method}, for nuanced finetuning, \textsl{ETHER}’s transformation strength seems to be too high to retain key semantic concepts in subject-driven generation, falling short in image quality with respect to other methods (e.g. also qualitatively depicted in Fig~\ref{fig:sample_perturbations}), despite achieving strong image diversity and text-prompt fidelity.

\begin{table}
    \centering
    \vspace{-0.2cm}
    \caption{\textbf{Subject-driven Generation Results.} We use $r$ to denote rank, and $n$ the number of diagonal blocks. We measure image quality (DINO, CLIP-I), text-prompt fidelity (CLIP-T) and image diversity (LPIPS). \textsl{ETHER+} addresses finegrained adaptation shortcomings of \textsl{ETHER} (c.f. Sec.~\ref{subsec:ether+}) and achieves strong performance with only few adaptation parameters.}
    \label{tab:dreambooth}
    \vspace{0.25cm}
    \resizebox{\linewidth}{!}{
    \begin{tabular}{lcccccc}
        \toprule
        \rowcolor{white}&\#params & DINO $\uparrow$ & CLIP-I $\uparrow$& CLIP-T $\uparrow$ & LPIPS $\uparrow$\\ 
        \midrule
        \rowcolor{white} Real Images           & -      & 0.703 & 0.864 & -    & 0.695\\ 
        DreamBooth                             & 859.5M & 0.644 & 0.793 & 0.236 & 0.709\\ 
        \rowcolor{lightgray}LoRA$_{r=4}$                           & 0.8M   & 0.660 & 0.796 & 0.231 & 0.714\\
        \rowcolor{lightgray}OFT$_{n=4}$         & 11.6M  & 0.652 & 0.794 & 0.241 & 0.725\\ 
        \rowcolor{lightblue}\textsl{ETHER}     & 0.1M   & 0.567 & 0.746 & \textbf{0.256} & \textbf{0.766}\\
        \rowcolor{lightgreen}\textsl{ETHER+} & 0.4M & \textbf{0.666} & \textbf{0.800} & 0.240 & 0.729\\ 
        \bottomrule
    \end{tabular}
    }
    \vspace{-0.3cm}
\end{table}

\begin{table}%[t]
    \centering
    \vspace{-0.3cm}
    \caption{\textbf{Semantic Map to Image Results.} We use $n$ to denote the number of diagonal blocks. \textsl{ETHER} and particularly \textsl{ETHER+} achieve strong synthesis control (mIoU, Acc) with few parameters while retaining good image alignment (FID). We indicate with (+ magn. r.f.) the OFT version with magnitude re-fitting.}
    \label{tab:s2i}
    \vspace{0.25cm}
    \resizebox{0.9\linewidth}{!}{
    \begin{tabular}{lcccc}
        \toprule
          &\#params & mIoU $\uparrow$ & Acc $\uparrow$ & FID $\downarrow$\\
        \midrule
        Encoder-only                        & 0     & 8.2   & 38.0  & 41.2 \\
        \rowcolor{lightgray}OFT$_{n=4}$        & 13.2M & 24.5  & 62.8 & 31.1\\
        \rowcolor{lightgray}OFT$_{n=4}$ (+ magn. r.f.)  & 13.4M & 24.6  & 63.3 & \textbf{30.8}\\
        \rowcolor{lightblue}\textsl{ETHER}  & 0.1M  &  24.6 & 63.3 & 32.0\\ 
        \rowcolor{lightgreen}\textsl{ETHER+} & 0.4M  & \textbf{27.3} & \textbf{68.1}& 31.0\\
        \bottomrule
    \end{tabular}
    }
    \vspace{-0.5cm}
\end{table}

\subsubsection{Controllable Image Generation}\label{subsubsec:controllablegen}
This section applies \textsl{ETHER} for controllability of Stable Diffusion following \citet{qiu2023oft} for the Semantic Map to Image (S2I) task on ADE20K \citep{zhou_semantic_2018}. We use the trainable encoder from ControlNet \newcite{zhang_adding_2023} for the control signal and perform finetuning on the Stable Diffusion weights only. We report a baseline with just the control signal encoder to highlight relative gains through finetuning. 
Evaluations are performed on 2000 images generated from the validation set using mean Intersection-over-Union (mIoU) and accuracy of semantic maps over generated images using UperNet-101~\citep{xiao2018unified} pretrained on ADE20K. Finally, we measure the similarity between generated and original images via FID~\citep{heusel2018gans}. For OFT, we also test magnitude re-fitting \citep{qiu2023oft} for an additional epoch.

\textit{Quantitative Results.} Results are depicted in Tab.~\ref{tab:s2i}, and clearly demonstrate competitive control with both \textsl{ETHER} and \textsl{ETHER+}. Unlike subject-driven image generation, we find that \textsl{ETHER} performs on the same level as OFT multiplicative finetuning while using over $100\times$ fewer parameters (e.g. $24.6$ versus $24.5$ mIoU of OFT with $0.1M$ versus $13.2M$ parameters). Introducing magnitude re-fitting to OFT yields only limited gains while adding $0.2M$ parameters.
Similar to Tab.~\ref{tab:dreambooth} for subject-driven image generation, we find that for controllable image synthesis, the \textsl{ETHER+} relaxation provides additional performance gains (e.g. $27.3$ vs $24.5$ mIoU and $68.1$ vs $62.8$ Acc against OFT). Taking into account the more robust (Fig.~\ref{fig:miou_vs_lr}) and faster convergence (Fig.~\ref{fig:convergence_cn}), this presents \textsl{ETHER+} as a practically attractive finetuning alternative.

\subsection{\textsl{ETHER} for Language Models Adaptation}\label{subsec:language_models_adaptation}
To understand the applicability of the \textsl{ETHER} transformation family in the language domain, we follow \citet{liu2023parameterefficient}'s and \citet{hu2022lora}'s experimental setup.
For fair comparisons, we run grid searches over the most relevant hyperparameters in common value ranges. For additional implementation details, please refer to App.~\ref{supp:exp_details}.

\subsubsection{Natural Language Understanding}\label{subsubsec:langunderstand}
We begin by deploying \textsl{ETHER} and \textsl{ETHER+} on the widely utilized \citep{devlin2019bert,liu_roberta_2019,he2023debertav,kopiczko_vera_2023} GLUE benchmark \citep{wang2018glue}, finetuning a pretrained DeBERTaV3-base model \citep{he2023debertav} following \citet{liu2023parameterefficient}, from which we report the baselines' results. GLUE comprises various English sentence understanding tasks, such as inference tasks (MNLI, QNLI, RTE), classification of sentiment (SST-2) or correct English grammatical structures (CoLA), and semantic similarity and equivalence prediction (MRPC, QQP, STS-B). CoLA scores report the Matthews correlation coefficient, MNLI matched accuracy, and STS-B average correlation. All other tasks are evaluated on accuracy. 

\textit{Quantitative Results.} Results in Tab.~\ref{tab:glue} show that \textsl{ETHER} and \textsl{ETHER+} match and even outperform previous methods with significantly fewer parameters. For example, \textsl{ETHER} outperforms the second-best BOFT on the RTE inference task ($89.53$ vs $88.81$) or equivalence prediction on MRPC ($93.68$ vs $92.40$) while using just one-ninth of the parameters ($0.085M$ compared to $0.75M$). \textsl{ETHER+} sets both the best performance on STS-B and particularly the highest overall score ($90.10$) using less than half of the parameters of BOFT.
These results provide additional support for the practical viability of \textsl{ETHER} transformations, now for natural language adaptation - being a strong, but much more parameter-efficient competitor.

\addtolength{\tabcolsep}{-4pt} 
\begin{table}[t]
    \centering
    \caption{\textbf{GLUE benchmark.} Comparisons of different methods finetuning DeBERTaV3-base. Results of all baselines are taken from \cite{liu2023parameterefficient}. We use $r$ to denote rank, and $n$ the number of diagonal blocks. As can be seen, \textsl{ETHER} and \textsl{ETHER+} achieve competitive performances across metrics while utilizing fewer parameters (up to a magnitude in the case of \textsl{ETHER}) while also retaining all practical benefits such as learning rate robustness depicted e.g. in Sec.~\ref{sec:intrigue}.}
    \label{tab:glue}
    \vspace{0.25cm}
    \resizebox{1\linewidth}{!}{
    \begin{tabular}{lccccccccc|c}
           \toprule
        &\#params & MNLI$\uparrow$& SST-2$\uparrow$ & CoLA$\uparrow$ & QQP$\uparrow$ & QNLI$\uparrow$ & RTE$\uparrow$ & MRPC$\uparrow$ & STS-B$\uparrow$ & \textbf{Avg}$\uparrow$\\ 
        \midrule
        Full Finet.                             &184M &89.90&95.63&69.19&\textbf{92.40}&94.03&83.75&89.46&91.60&88.25\\
        \rowcolor{lightgray}BitFit   & 0.10M&89.37&94.84&66.96&88.41&92.24&78.70&87.75&91.35&86.20\\
        \rowcolor{lightgray}H-Adapter& 1.22M&90.13&95.53&68.64&91.91&94.11&84.48&89.95&91.48&88.28\\
        \rowcolor{lightgray}P-Adapter& 1.18M&90.33&95.61&68.77&92.04&94.29&85.20&89.46&91.54&88.41\\
        \rowcolor{lightgray}LoRA$_{r=8}$            &1.33M&90.65&94.95&69.82&91.99&93.87&85.20&89.95&91.60&88.50\\
        \rowcolor{lightgray}AdaLoRA            &1.27M&\textbf{90.76}&96.10&71.45&92.23&\textbf{94.55}&88.09&90.69&91.84&89.46\\
        \rowcolor{lightgray}OFT$_{n=16}$        &0.79M&90.33&96.33&\textbf{73.91}&92.10&94.07&87.36&92.16&91.91&89.77\\
        \rowcolor{lightgray}BOFT$_{n=8}^{m=2}$  &0.75M&90.25&\textbf{96.44}&72.95&92.10&94.23&88.81&92.40&91.92&89.89\\
        \rowcolor{lightblue}\textsl{ETHER}  &0.09M &90.23  &96.10 &71.31 &91.42 &94.31 &\textbf{89.53} &\textbf{93.68} &92.30&89.86\\
        \rowcolor{lightgreen}\textsl{ETHER+} &0.33M& 90.52 &96.33 &72.64 &92.22 &94.33 &\textbf{89.53} &92.89 & \textbf{92.35} & \textbf{90.10}\\  
        \bottomrule
    \end{tabular}}
\vspace{-0.1cm}
\end{table}
\addtolength{\tabcolsep}{4pt}

\subsubsection{Instruction Tuning}\label{subsubsec:instruction_tuning}

Our instruction tuning experiments make use of Llama-2-7B \newcite{Touvron2023Llama2O} as pretrained model, finetuning it on the Alpaca dataset \citep{alpaca} for one epoch. To operate on a consumer GPU, we truncate the maximum sequence length to 256 and use bfloat16 precision \citep{kalamkar2019study}.
We evaluate 0-shot performance of our instruction-tuned model on (i) Massive Multitask Language Understanding (MMLU) \citep{hendrycks_measuring_2021} with 57 different tasks in four different subjects (STEM, Humanities, Social Sciences, Others); (ii) the AI2 Reasoning Challenge (ARC) \cite{clark_think_2018}, a common-sense reasoning dataset of questions from science grade exams; (iii) TruthfulQA \citep{lin2022truthfulqa} comprising 817 questions spanning 38 categories testing how much the model (wrongly) relies on imitation of human text to answer. 

\textit{Quantitative Results.} Results in Tab.~\ref{tab:lm} show that both \textsl{ETHER} and \textsl{ETHER+} outperform comparable finetuning approaches while utilizing fewer parameters. Across all metrics, the Llama-2-7B baseline is consistently surpassed by significant margins (e.g. $44.87$ MMLU for \textsl{ETHER+} vs the $41.81$ baseline, or $46.50$ vs $42.92$ ARC score).
Despite being the most parameter-efficient method, \textsl{ETHER} outperforms all baselines with comparable number of parameters, such as the recently introduced VeRA \citep{kopiczko_vera_2023} with rank $r=64$, and LoRA rank $1$. Surprisingly, increasing the rank of VeRA to $256$ leads to a decrease in performance, while LoRA rank $8$ shows better results but is still outperformed on MMLU despite having $16\times$ more parameters.
On the other hand, \textsl{ETHER+} surpasses all other methods across all benchmarks, while having $4\times$ fewer parameters than LoRA rank $8$.

\begin{table}[t]
    \centering
    \caption{\textbf{Instruction Tuning.} We use $r$ to denote rank, and $n$ the number of diagonal blocks. 
    Both \textsl{ETHER} and \textsl{ETHER+} outperform LoRA/OFT which use up to a magnitude more parameters, and beat VeRA with similar parameter counts.}
    \label{tab:lm}
    \vspace{0.25cm}
    \resizebox{0.85\linewidth}{!}{
    \begin{tabular}{lccccc}
        \toprule
        &\#params & MMLU$\uparrow$& ARC$\uparrow$ & Tru-1$\uparrow$ & Tru-2$\uparrow$\\ 
        \midrule
        Llama-2-7B                                         &  -    & 41.81 & 42.92 & 25.21& 38.95\\       
        \rowcolor{lightgray}VeRA$_{r=64}$                 & 0.27M  & 42.30 & 45.13 & 27.41 & 41.04\\
        \rowcolor{lightgray}VeRA$_{r=256}$                 & 1.05M  & 42.21 & 43.85 & 25.33 & 39.02\\
        \rowcolor{lightgray}LoRA$_{r=1}$                  & 0.52M  & 42.40 & 44.62 & 27.05 & 41.94\\
        \rowcolor{lightgray}LoRA$_{r=8}$                  & 4.19M  & 43.61 & 46.16 & 28.76 & 42.21\\
        \rowcolor{lightgray}OFT$_{n=256}$                & 2.09M & 42.92 & 44.88 & 27.42 & 41.11\\
        \rowcolor{lightblue}\textsl{ETHER}$_{n=32}$ & 0.26M & 44.57 & 45.14 & 27.91 & 41.83\\
        \rowcolor{lightgreen}\textsl{ETHER+}$_{n=32}$ & 1.04M & \textbf{44.87} & \textbf{46.50} & \textbf{29.38} & \textbf{43.51}\\     
        \bottomrule
    \end{tabular}}
\vspace{-0.1cm}
\end{table}

\subsection{Hyperspherical Energy for Effective PEFT}
\label{subsec:orthogonality}

\begin{figure}[t]
  \centering
  \vspace{0.2cm}
   \includegraphics[width=\linewidth]{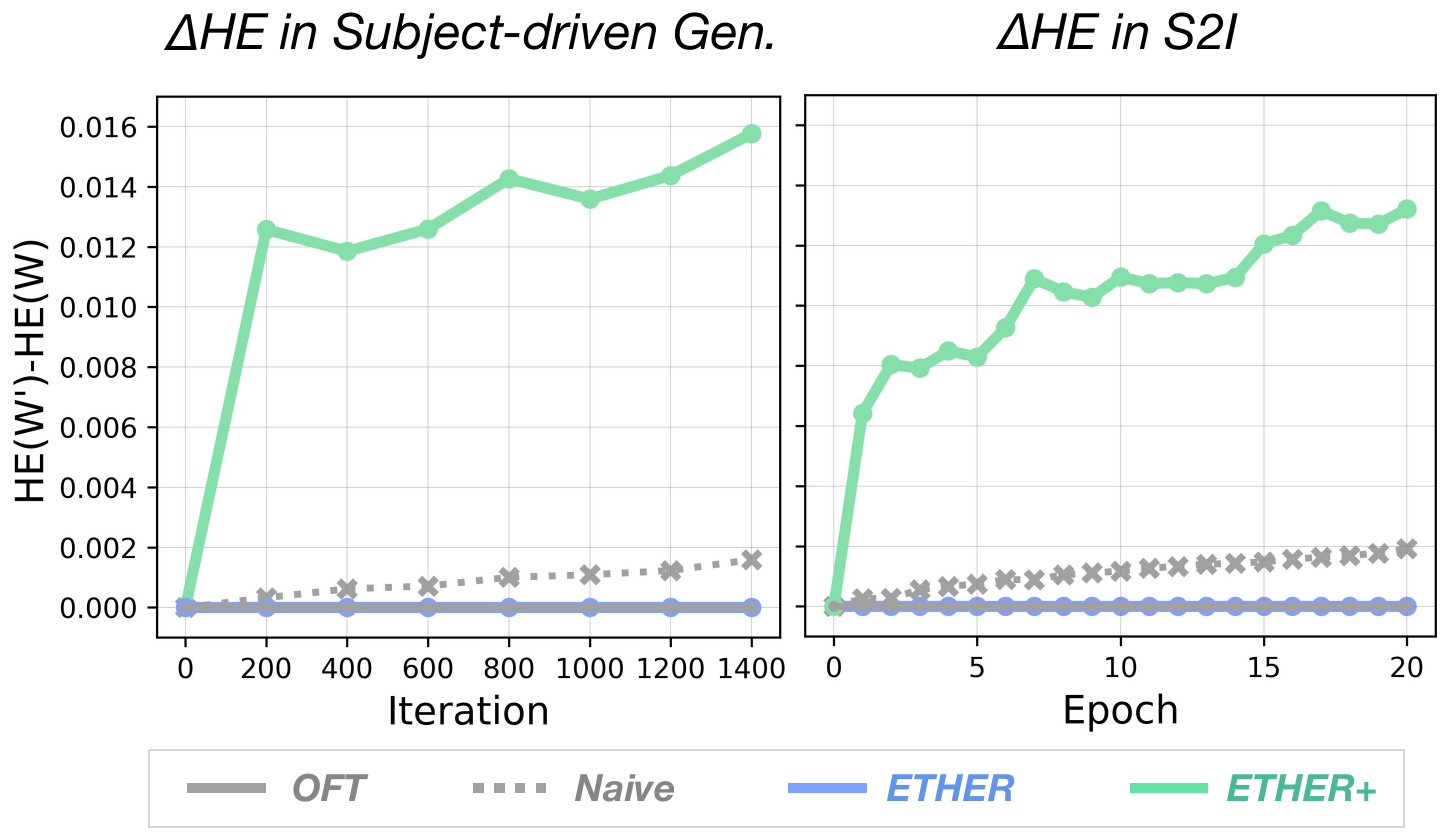}
   \vspace{-0.4cm}
   \caption{\textbf{Difference in HE} between finetuned/pretrained models for Subject-driven Generation and S2I. Notice that by removing the orthogonality constraint, both \textsl{ETHER+} and Naive alter the HE of the pretrained model, while OFT and \textsl{ETHER} do not.}
   \label{fig:he_impact}
\vspace{-0.1cm}
\end{figure}

\citet{qiu2023oft} link finetuning stability and performance obtained by transforming the weights via matrix-multiplication to the orthogonality of the transformations, and a consequently unaltered hyperspherical energy (HE). % 
To test this assumption, we have included an OFT control baseline (\textit{Naive}), which does not utilize orthogonality constraints, on the same finetuning settings in which OFT was proposed. Results at convergence, as reported in Tab.~\ref{tab:he_tab}, do not show significant differences, while actually introducing the overhead of computing the Cayley parametrizations (which also involve computing the inverse of a matrix). We also included the \textit{Naive} baseline in the learning rate robustness studies in Fig. \ref{fig:distance_vs_lr} and Fig. \ref{fig:miou_vs_lr}, showcasing that while differences are present for high learning rates, the optimal working range remains unaltered. Finally, we validate that the HE indeed varies during training, as reported in Fig. \ref{fig:he_impact}. 

In contrast, on these same evaluations, our newly proposed \textsl{ETHER} transformation family, by introducing a boundary on the Euclidean distance on the transformation side, achieves stronger performance and greater robustness. This is especially true for the non-orthogonal \textsl{ETHER+}, which alters the overall HE even more than \textit{Naive} (Fig. \ref{fig:he_impact}).
This evidence diminishes the role of the HE and instead emphasizes the greater importance of the Euclidean distance, establishing the \textsl{ETHER} family as a favorable option in multiplicative finetuning settings.

\begin{table}
\caption{\textbf{OFT vs Naive.} OFT performance-test against its non-orthogonal counterpart Naive. We show that results don't differ significantly, questioning the relevance of HE retaining for finetuning performance.
}
\vspace{0.1cm}
\resizebox{1\linewidth}{!}{
\begin{tabular}{l|cccc|ccc}
\multirow{2}{*}{} & \multicolumn{4}{c|}{\textbf{Subject-driven Generation}} & \multicolumn{3}{c}{\textbf{S2I}} \\ 
& DINO      & CLIP-I & CLIP-T & LPIPS & mIoU & Acc & FID \\ 
 \toprule
\rowcolor{lightgray} OFT$_{n=4}$ & 0.652 & 0.794 & 0.241 & 0.725 & 24.5 & 62.8 & 31.1\\
 \rowcolor{lightgray}Naive$_{n=4}$ & 0.648 & 0.793 & 0.245 & 0.730 & 24.3 & 62.9 & 29.9\\
\end{tabular}
}
\vspace{-0.2cm}
\label{tab:he_tab}
\end{table}

   %sec/4_experiments}
\section{Conclusions}
\label{sec:5_conclusions}
Our paper introduces the \textsl{ETHER} family of transformations for parameter-efficient finetuning. Based on the Householder formulation of hyperplane reflections, \textsl{ETHER} methods frame finetuning as a search for unit normal vectors that define hyperplanes along which weight vectors are reflected. 
In doing so, \textsl{ETHER} (and its relaxation \textsl{ETHER+} for more finegrained adaptation) fix (or upper bound) the distance of learned transformations from the identity matrix (the transformation neutral element), thereby minimizing the risk of finetuning divergence.
Put together, \textsl{ETHER} methods operate more parameter-efficiently than other PEFT methods (e.g., around 10-100 times less than LoRA or OFT), have higher learning rate robustness and encourage fast convergence. Consequently, \textsl{ETHER} transformations require less expansive hyperparameter searches to achieve good performance, making them very attractive for practical deployment.

\textbf{Limitations.} Of course, there is no free lunch. While both \textsl{ETHER} and its relaxation \textsl{ETHER+} show strong results with few parameters across a broad range of tasks, increasing the expressive power of the transformation is not as straightforward as in other methods, such as LoRA, where one can adjust the rank parameter to more closely approximate full finetuning.
Moreover, multiplicative methods introduce a computational overhead during training compared to additive methods.
Thanks to our block-parallel scheme, 
we make significant progress towards closing the gap between multiplicative and additive approaches; however, multiplicative methods still lag behind. This introduces a trade-off between parameter efficiency and computational overhead when achieving similar performance levels.

%%%%%%%%%%%%%%%%%%%%%%%%%%%%%%%%%%%%%%%%%%
\section*{Impact Statement} 
This paper presents work that looks into better and more efficient finetuning of foundation models. By bringing down the need for compute-expensive hyperparameter grid searches and encouraging fast convergence, both the cost and environmental footprint of serving individually adapted models at scale can be brought down.
Of course, with most advancement in the field of Machine Learning, there is potential for misuse and societal consequences, however, none of which we feel are specific to our proposed method and which need to be highlighted explicitly.

\section*{Acknowledgements}
Massimo Bini was supported by Bosch Industry on Campus Lab at University of Tübingen.
Karsten Roth thanks the European Laboratory for Learning and Intelligent Systems (ELLIS) PhD program and the International Max Planck Research School for Intelligent Systems (IMPRS-IS) for support.
Zeynep Akata and Karsten Roth were supported by DFG project number 276693517, by BMBF FKZ: 01IS18039A, by the ERC (853489 - DEXIM), by EXC number 2064/1 – project number 390727645.
   %sec/5_conclusions}

\bibliography{main}
\bibliographystyle{icml2024}

%%%%%%%%%%%%%%%%%%%%%%%%%%%%%%%%%%%%%%%%%%%%%%%%%%%%%%%%%%%%%%%%%%%%%%%%%%%%%%%
%%%%%%%%%%%%%%%%%%%%%%%%%%%%%%%%%%%%%%%%%%%%%%%%%%%%%%%%%%%%%%%%%%%%%%%%%%%%%%%
% APPENDIX
%%%%%%%%%%%%%%%%%%%%%%%%%%%%%%%%%%%%%%%%%%%%%%%%%%%%%%%%%%%%%%%%%%%%%%%%%%%%%%%
%%%%%%%%%%%%%%%%%%%%%%%%%%%%%%%%%%%%%%%%%%%%%%%%%%%%%%%%%%%%%%%%%%%%%%%%%%%%%%%
\newpage
\appendix
\onecolumn
\clearpage
\setcounter{page}{1}
\noindent

\normalsize
\begin{center}
\Large{\textbf{Appendix}}   
\end{center}
\normalsize

In this appendix, we augment the main paper with additional, qualitative evidence for the learning rate robustness of \textsl{ETHER} transformations in \cref{supp:11_qualit_robustness}. In addition, we also provide benchmark-specific qualitative examples for subject-driven and controllable image generation in \cref{supp:12_qualit}. For all experiments - both those in the main paper and supplementary results, we then list all relevant details in \cref{supp:exp_details} for our studies on finetuning in subject-driven image generation (\S\ref{supp:subj_gen}), controllable image synthesis (\S\ref{supp:contr_gen}), natural language understanding tasks (\S\ref{supp:nlu_gen}) and instruction tuning (\S\ref{supp:inst_tun}).
We then provide two additional \textsl{ETHER} ablations in \cref{suppsec:ablations} - for the number of block-diagonals and the specific double-sided application in \textsl{ETHER+}. Finally, we present preliminary results on the Visual Task Adaptation Benchmark (\S\ref{suppsec:vtab}).

%%%%%%% [A] Learning Rate Robustness %%%%%%%%%%
\section{Qualitative Evidence of Learning Rate Robustness} \label{supp:11_qualit_robustness}
As introduced in Sec.~\ref{sec:3_method}, when finetuning with \textsl{ETHER} transformation, by construction, the learning rate only controls the speed with which reflection angels change. 
As a consequence, \textsl{ETHER} methods are much more robust to learning rate choices, and less likely to diverge and cause model deterioration. This allows for user control over the convergence speed while minimizing the risk of model collapse during training.
To demonstrate this, Sec.~\ref{sec:intrigue} introduced both a qualitative example comparing the impact of minimal and maximal perturbation strength on the model output in Fig.~\ref{fig:sample_perturbations}, and quantitative evaluations on the Semantic Map to Image task against learning rate choices in Figs.~\ref{fig:miou_vs_lr} and ~\ref{fig:convergence_cn}.

\begin{figure}[H]
  \centering
  \makebox[\textwidth]{
\includegraphics[width=1.02\textwidth]{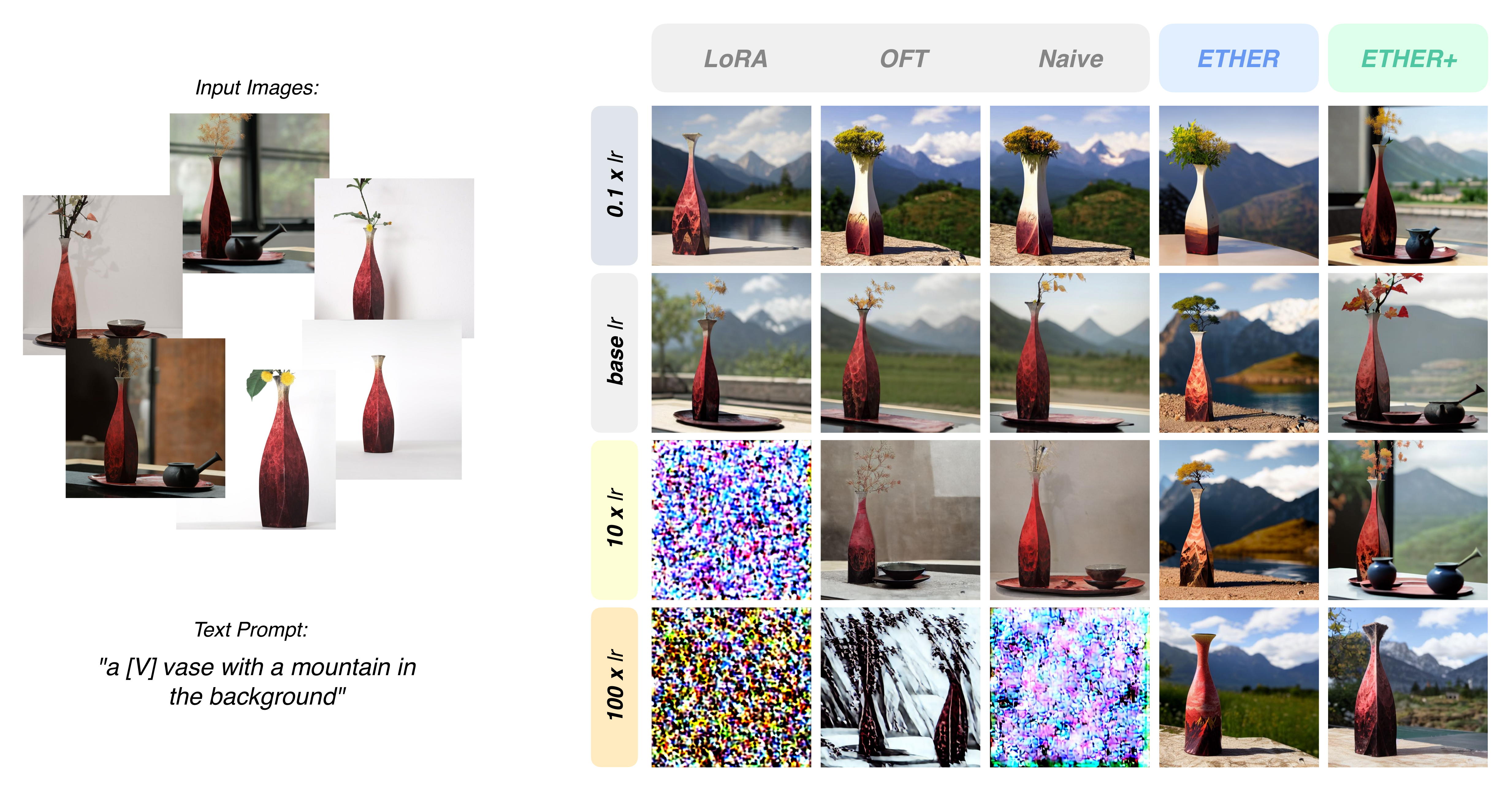}
}
\caption{Qualitative visualization of learning rate robustness of \textsl{ETHER} and \textsl{ETHER+} in subject-driven generation finetuning. 
We see how \textsl{ETHER} methods are able to consistently produce good results avoiding model deterioration. Specifically, \textsl{ETHER+} shows impressive capabilities, being able to follow the subject-prompt instructions in the widest learning rate range.}

\label{suppfig:rate_db}
\end{figure}

In this section, we augment Sec.~\ref{sec:intrigue} and provide additional qualitative results and impressions to highlight the non-deteriorating nature of \textsl{ETHER} transformation. For this, we showcase subject-driven generation results using different finetuning methods in Fig.~\ref{suppfig:rate_db}, with default generations using the best learning rate. We then systematically increase the finetuning learning rate by $10$ and by $100$ times, and visualize the correspondingly generated output. 
As can be seen, for $10\times$ higher learning rates OFT and Naive fail to follow the text prompt, while LoRA finetuning quickly collapses. With $10\times$ lower learning rates instead, OFT, Naive and \textsl{ETHER} are not able to generate the subject correctly in the predefined number of iterations.

\newpage

%%%%%%% [B] Qualitative Examples %%%%%%%%%%
\section{Qualitative Examples for \textsl{ETHER} Finetuning} 
\label{supp:12_qualit}

We show some qualitative results by using the finetuning methods proposed in this paper.

\subsection{Subject-driven Generation.}
\begin{figure}[H]
  \centering
  \makebox[\textwidth]{
\includegraphics[width=.95\textwidth]{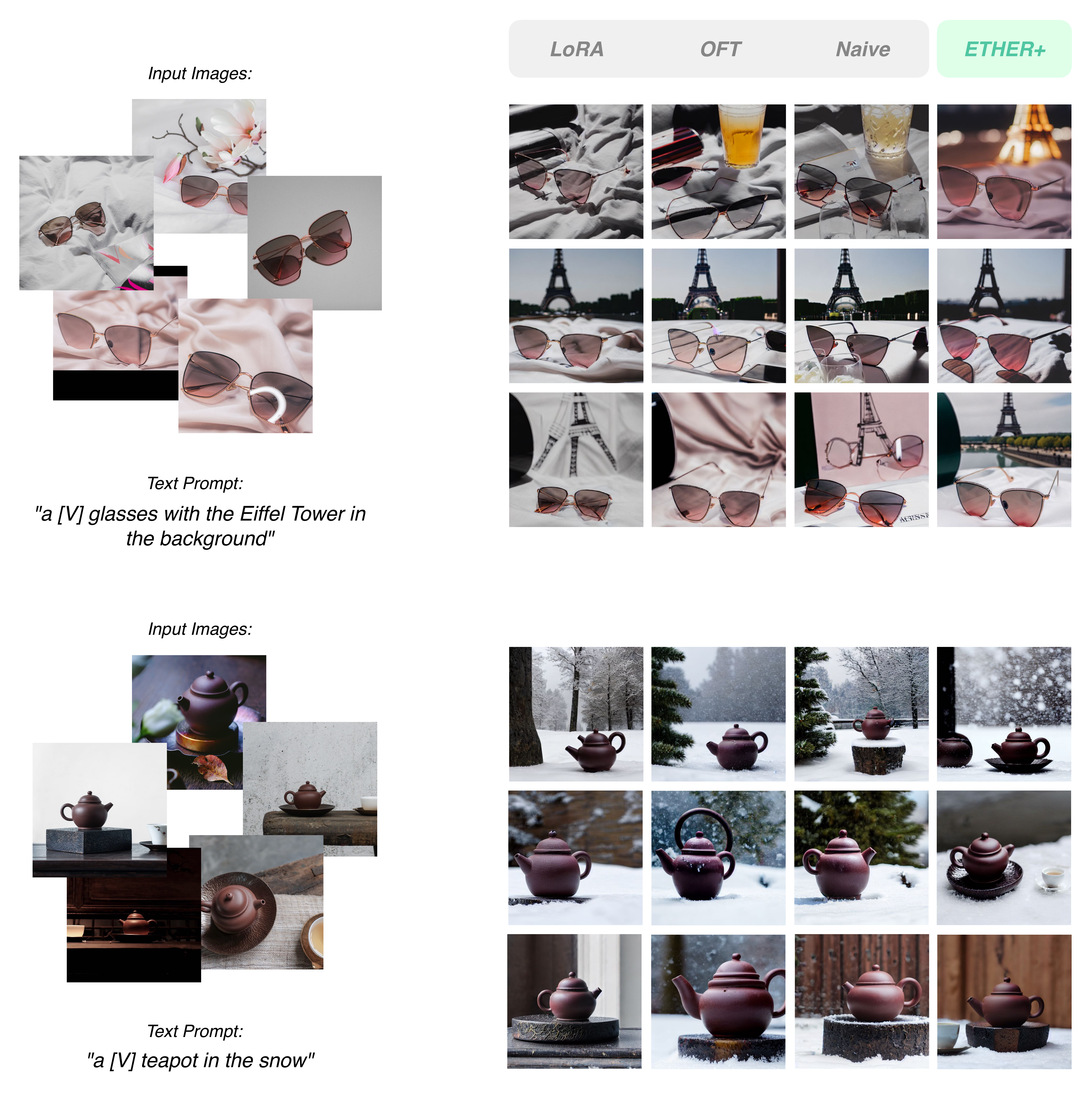}
}
\vspace{-0.1cm}
\caption{Subject-driven Generation results. Each row shares initial latent noise (notice row-wise similarities). We can see that \textsl{ETHER+} method is better at adapting the model to the subjects. Notice how for the pink sunglasses, OFT and Naive fail in following the prompt.}
\label{fig:results_db}
\end{figure}

In Figure \ref{fig:results_db} we report subject-driven generation examples. In particular, for a fair comparison, we report images which come from the same noise vector in the Stable Diffusion latent space.  For the \textit{sunglasses} images, we see how non-\textsl{ETHER} methods manage to reproduce the subject, but fail to follow the text prompt in most cases. Interestingly in the first row, we notice how \textsl{ETHER+} is able to properly control the generation, by transforming the yellow area (associated to a beer in other models) in an enlightened Eiffel Tower. For the \textit{teapot} images instead, we see how \textsl{ETHER+} is able to better keep the appearances of the subject.
\vspace{-5pt}

\subsection{Controllable Generation.}
\begin{figure}[H]
  \centering
  \makebox[\textwidth]{
\includegraphics[width=1.02\textwidth]{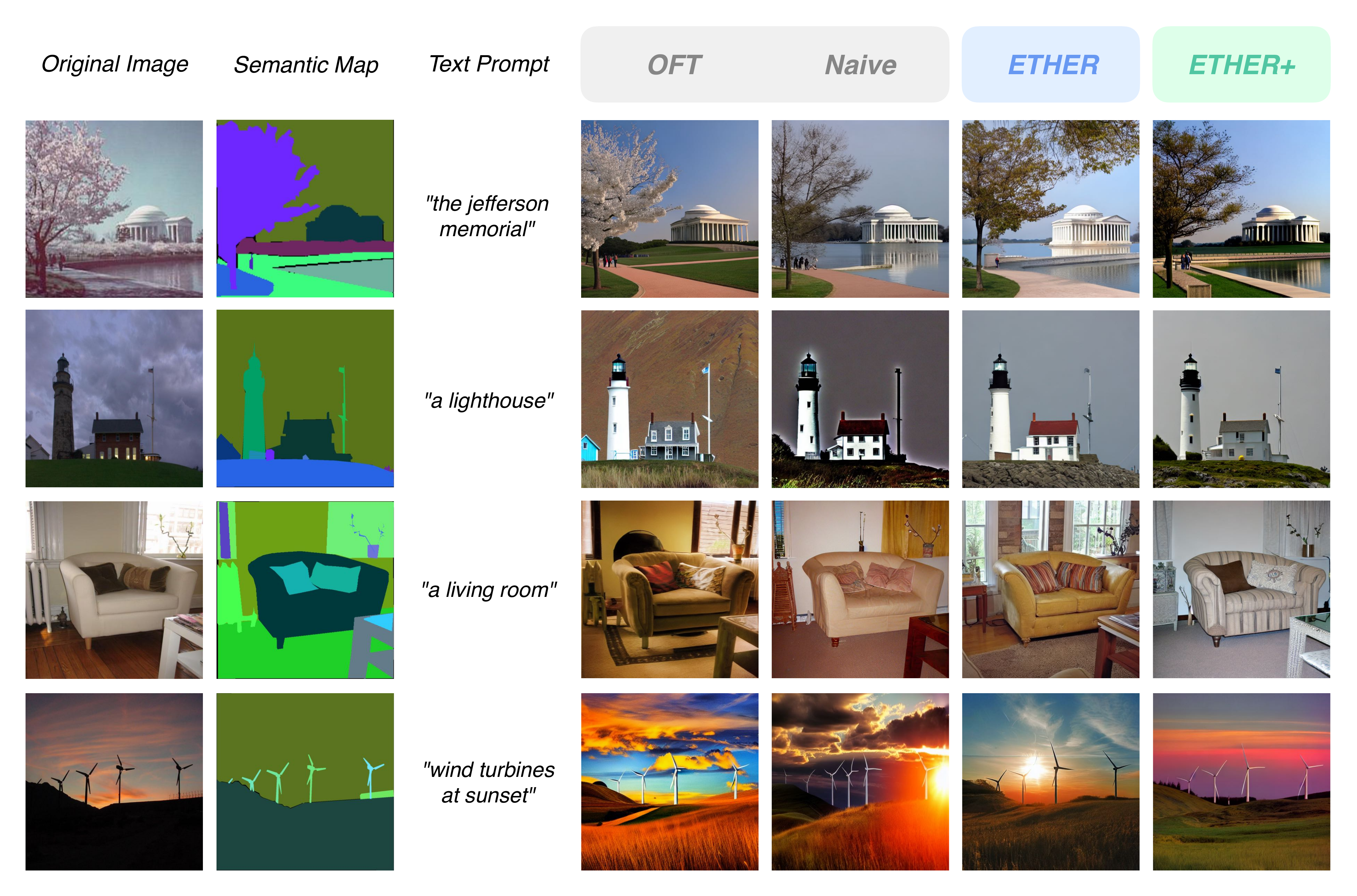}
}
\vspace{-0.4cm}
\caption{Semantic Map to Image Qualitative Results. We notice how in the first row all models but \textsl{ETHER+} fail to control the image correctly. Overall \textsl{ETHER+} controlled images show better control.}
\label{fig:results_cn}
\end{figure}

In Figure \ref{fig:results_cn} we show some examples from the Semantic Map to Image task. In particular, we notice how in the first row all models but \textsl{ETHER+} fail to control the image correctly, not being able to separate the land from the water.
Additionally, in the second row OFT fails to generate the sky, while Naive presents a halo effect. These examples showcase the abilities of \textsl{ETHER+} finetuning over the other methods.
To show broader controllable capabilities, we also report few qualitative examples with \textsl{ETHER} methods trained with Landmarks and Canny Edge Maps control signals on CelebA-HQ \cite{karras2018progressive} and COCO 2017 \cite{lin2015microsoft} datasets respectively.

\begin{figure}[H]
  \centering
  \makebox[\textwidth]{
\includegraphics[width=1.02\textwidth]{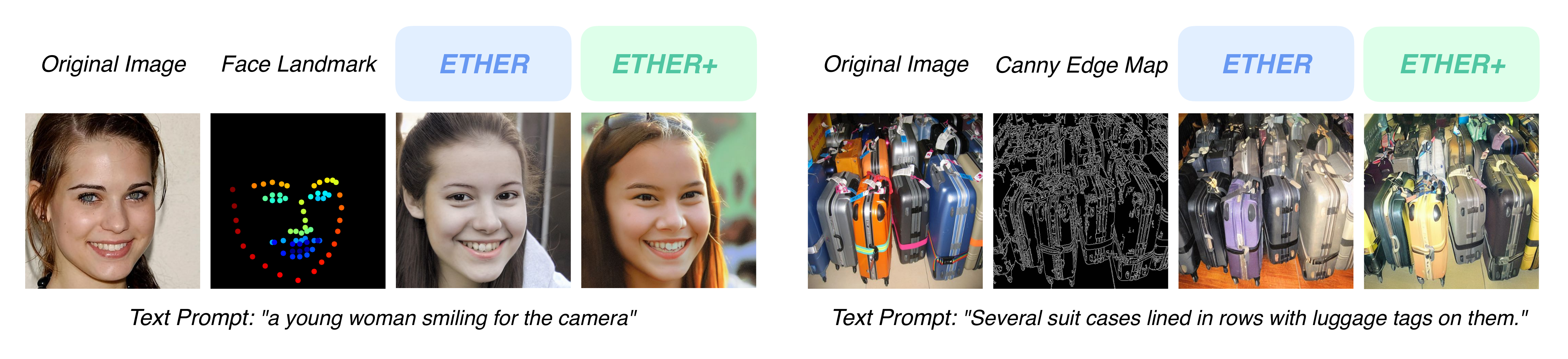}
}
\vspace{-0.4cm}
\caption{Examples of Landmark to Face (left) and Canny Edge Map to Image (right) controlled generation with \textsl{ETHER} methods.}
   \label{fig:results_cn_lmk}
\end{figure}

%%%%%%% [C] Experimental Details %%%%%%%%%%
\section{Experimental Details}\label{supp:exp_details}
This section provides additional experimental details for replication not listed in the main benchmark experimental section~\ref{sec:4_experiments}. 
It is worth noting that while in most of our experiments we do not employ regular dropout \citep{dropout}, \citet{liu2023parameterefficient} proposes a multiplicative dropout form specifically designed for multiplicative finetuning methods, which we did not test in this study. We hypothesize that this specialized dropout technique could potentially work better than regular dropout for \textsl{ETHER} and \textsl{ETHER+} as well.
We also note that \citet{qiu2023oft} report OFT's number of parameters as half of the actual trainable parameters due to the redundancy in the skew symmetric matrices $S^B$ in the Cayley parametrization of $Q^B$. Basically, we they report the storage parameters for $Q^B$ rather than the training parameters. For consistency and fair comparisons, we follow the same convention for OFT throughout our paper.

\subsection{Subject-driven Generation}\label{supp:subj_gen}
For subject-driven generation, we follow the same setting listed in DreamBooth \cite{ruiz_dreambooth_2023}, using DreamBooth and OFT \cite{qiu2023oft} baselines as implemented in official \href{https://github.com/Zeju1997/oft}{OFT} GitHub repository. The additional trainable layers follow \cite{qiu2023oft} and are added to the Q,K,V layers and the projection layer inside every attention module.
The training is performed over 1400 iterations for each method, evaluating the generation results every 200 iterations at selecting the best one (typically around 1200 iterations). 
For DreamBooth and OFT, we follow the original implementations and use a learning rate of $5\times 10^{-6}$ and $6\times 10^{-5}$ respectively, with a batch size of $1$. For Naive - the non-orthogonal OFT variant - we use the same setting of OFT for a fair comparison. For LoRA we select a learning rate of $6\times 10^{-4}$. For \textsl{ETHER} and \textsl{ETHER+}, we use a learning rate of $6\times 10^{-3}$. We perform the training on a Tesla V100-32GB GPU.

\subsection{Controllable Generation}\label{supp:contr_gen}
For our experiments on controllable image generation we follow the setting of \citet{qiu2023oft}, using the signal encoder from ControlNet \citep{zhang2023controlnet} (comprising 8 trainable convolutional layers, accounting for 3.1M additional learnable parameters). Finetuning parameters are added to the Q,K,V layers as well as the projection layer of the attention modules and the subsequent feedforward layers. 
As baselines, we use the official implementation of \href{https://github.com/Zeju1997/oft}{OFT}. Similarly to \citet{qiu2023oft}, for OFT and Naive we use a learning rate of $1\times 10^{-5}$. For \textsl{ETHER} and \textsl{ETHER+} we use a larger learning rate of $1\times 10^{-3}$. For all experiments, we upper bound the learning rate of the signal encoder to $1\times 10^{-4}$. We perform all the training runs on a single Nvidia-A100-40GB with a batch size of 10. As listed in Sec.~\ref{subsubsec:controllablegen} and expanded in Sec.~\ref{suppsec:lora}, we tried to utilize LoRA for controllable generation as well but found no comparable results even after extensive trials with different hyperparameters.

\subsection{Natural Language Understanding}\label{supp:nlu_gen}
For our GLUE benchmark experiments finetuning DeBERTaV3-base \cite{he2023debertav}, we make use of the \href{https://github.com/huggingface/peft}{peft} Hugging Face repository \citep{peft} as the basis for our implementations.
To compare our results with those of \citet{liu2023parameterefficient}, we follow their implementation and apply \textsl{ETHER} and \textsl{ETHER+} to all the linear layers in every transformer block. The relevant hyperparameters for each task are reported in Tab.~\ref{tab:glue_params}. All training runs are conducted on a single Nvidia-A100-40GB GPU.

\begin{table*}[!h]
    \centering
    \caption{GLUE benchmark hyperparameters.}
    \label{tab:glue_params}
    \vspace{0.1cm}
    \scalebox{0.93}{
    \begin{tabular}{cccccccccc}
        \toprule
        Method         &  Hyperparameters& MNLI & SST-2 & CoLA & QQP & QNLI & RTE & MRPC& STS-B \\ 
        \midrule
                       &  Learning Rate  & 8e-4 & 1e-3  & 1e-3 &3e-4 &1e-3  & 1e-3 & 3e-4 & 2e-3 \\ 
                       &  Batch Size     & 32   & 32    & 32   &8    &8     &  32  &  32  & 8    \\ 
        \textsl{ETHER} &  Num. Epochs    & 9    & 14    & 10   &20   &7     &  13  &  14  & 8    \\ 
                       &  Dropout        & 1e-3 & 1e-3  & 1e-1 &1e-1 &1e-3  & 1e-2 & 1e-1 & 1e-1 \\ 
                       &  Max Seq. Len.  & 256  & 128   & 64   &320  &512   &  320 &  320 & 128  \\ 
        \midrule
                       &  Learning Rate  & 8e-4 & 1e-4  & 1e-3 &3e-3 &3e-3  & 3e-4 &8e-4  & 8e-4 \\ 
                       &  Batch Size     & 8    & 8     & 8    &32   &32    & 8    & 32   &  8   \\ 
        \textsl{ETHER+}&  Num. Epochs    & 8    & 10    & 6    &16   &5     & 35   & 17   &  11  \\ 
                       &  Dropout        & 1e-3 & 1e-3  & 1e-1 &1e-3 &1e-3  & 1e-3 & 1e-2 & 1e-3 \\ 
                       &  Max Seq. Len.  & 256  &  128  & 64   &320  &512   & 320  & 320  & 128  \\ 
        \bottomrule
    \end{tabular}}
\end{table*}

\subsection{Instruction Tuning}\label{supp:inst_tun}
For our Instruction Tuning experiments, we use the LoRA \cite{hu2022lora} finetuning implementation in the \href{https://github.com/Lightning-AI/lit-gpt}{lit-gpt} repository \cite{litgpt-2023} as baseline. For evaluations, we make use of \citet{eval-harness}'s benchmark implementations.
For the recently proposed VeRA \cite{kopiczko_vera_2023} baseline, we reproduce the model implementation following their best performing method as described in the paper: sampling random $A$ and $B$ matrices with uniform kaiming initialization scaled by the matrix dimension, and a learnable, non-zero diagonalized vector initialized as a vector of all zeros apart for one element equal to $0.1$. Same for OFT, for which we follow the implementation in the official repository \href{https://github.com/Zeju1997/oft}{oft}, selecting the number of block-diagonal matrices such that the overall number of parameters becomes comparable with \textsl{ETHER+} and LoRA rank 8.
For all experiments, we use a cosine annealing learning rate scheduler, no dropout, and 1000 warmup steps. For LoRA, VeRA, and OFT we use AdamW optimizer with a weight decay of 0.01, while for \textsl{ETHER} methods, given the normalization happening on the parameters, weight decay would have limited impact and thus we set it to 0. For LoRA and VeRA, we keep $\alpha$ fixed with respect to the learning rate by setting it equal to the rank. For all experiments, we conduct an extensive grid search over learning rates and batch sizes. 
For each combination, we perform the LLama-2-7B \cite{Touvron2023Llama2O} finetuning over Alpaca \cite{alpaca} for one epoch.
All training runs are conducted on a single Nvidia-A100-40GB GPU, but could also be run on a consumer NVIDIA GeForce-RTX-3090-24G GPU.
\begin{table*}[!h]
    \centering
    \caption{Instruction Tuning hyperparameters.}
    \label{tab:glue_params}
    \vspace{0.1cm}
    \scalebox{0.93}{
    \begin{tabular}{cccccccc}
        \toprule
                 & VeRA$_{r=64}$ &VeRA$_{r=256}$ &LoRA$_{r=1}$&LoRA$_{r=8}$&  OFT$_{n=256}$ & \textsl{ETHER}$_{n=32}$ & \textsl{ETHER+}$_{n=32}$ \\ 
        \midrule
        Learning Rate & 5e-3 & 1e-3 & 3e-3 &5e-4 &  5e-4 & 2e-3&5e-3 \\
        Batch Size    & 32 & 32   & 8 & 8  &  16    & 8   &16  \\
        \bottomrule
    \end{tabular}}
\end{table*}

%%%%%%% [D] ETHER Ablations  %%%%%%%%%%
\section{\textsl{ETHER} Ablations}\label{suppsec:ablations}
This section details additional ablation experiments on the impact of the block-diagonality degree on the final performance, as well as experimental support to the theoretical motivation in Sec.~\ref{subsec:ether+} to apply the relaxed Householder transformation on both the left and right side of the weight matrix.

\subsection{Block-diagonal \textsl{ETHER} Performances} 
In \cref{tab:block_s2i} and \cref{tab:block_lm}, we compare the usage of multiple diagonal blocks for \textsl{ETHER} finetuning to allow for fast performance, especially in large models domain. Both tables augment our method description in Sec.~\ref{subsec:blockparallel} and the shortened results in Tab.~\ref{tab:load}.
In all cases, we notice that performance remains almost unaffected by the choice of block number, while on the other hand, the computational efficiency consistently increases ($8.22$ TFLOPs for $n=32$ versus $25.26$ TFLOPs for $n=1$ for Llama-2-7B). It is worth noting that results for \textsl{ETHER+} with ${n=32}$ show better performance with respect to less diagonalized counterparts.

\begin{table}[H]
    \centering
    \caption{Semantic Map to Image (S2I) results for different number of diagonal blocks $n$ on \textsl{ETHER} finetuning at epoch 10}
    \label{tab:block_s2i}
    \vspace{0.1cm}
    \scalebox{0.93}{
    \begin{tabular}{lcccc}
        \toprule
        \textsl{ETHER} &\#params & mIoU $\uparrow$ & Acc $\uparrow$ & FID$\downarrow$\\
        \midrule
        ${n=1}$   & 0.1M & 23.1 & 61.23  & 31.7\\
        ${n=4}$   & 0.1M & 22.9 & 60.92  & 30.5 \\
        ${n=16}$  & 0.1M & 22.3 & 60.35  & 30.7\\
        \bottomrule
    \end{tabular}}
\end{table}

\begin{table}[H]
    \centering
    \caption{Instruction Tuning results for different number of diagonal blocks $n$ on \textsl{ETHER} finetuning}
    \label{tab:block_lm}
    \vspace{0.1cm}
    \scalebox{0.93}{
    \begin{tabular}{lccccccc}
        \toprule
        \textsl{ETHER+}&\#params & TFLOPs & MMLU $\uparrow$& ARC $\uparrow$ & Tru-1$\uparrow$ & Tru-2 $\uparrow$\\ 
        \midrule
        %%%%%%%%%%%%% HIGH PARAM COUNT
        ${n=1}$   & 1.04M & 51.65 & 43.75 & 46.76 & 28.03 & 41.06\\
         ${n=4}$   & 1.04M & 18.66 & 43.91 & 45.73 & 27.54& 40.46\\
        ${n=32}$  & 1.04M & 9.04  & 44.87 & 46.50 & 29.38 & 43.51\\ 
        \bottomrule
    \end{tabular}}
\end{table}

\subsection{Double-sided Application of \textsl{ETHER+}} 

Finally, we provide a brief ablation study in Tab.~\ref{tab:dreambooth_abl}, comparing the \textsl{ETHER+} performance when applying the relaxed Householder transformations $H^+$ on only one side versus both sides. Although the parameter count doubles, we observe a significant increase in performance (e.g. $0.666$ vs $0.618$ in DINO score) as higher transformation distances can be achieved.

\vspace{-0.3cm}
\begin{table}[H]
    \centering
    \caption{Subject-driven Generation image quality results comparison (at iteration 1200) among standard \textsl{ETHER+} and its version only applied on one side of the weight matrix.}
    \label{tab:dreambooth_abl}
    \vspace{0.1cm}
    \scalebox{0.93}{
    \begin{tabular}{lcccc}
        \toprule
        &\#params & DINO $\uparrow$ & CLIP-I $\uparrow$\\
        \midrule
        \textsl{ETHER+ (one-sided)} & 0.2M & 0.618 & 0.777 \\
        \textsl{ETHER+} & 0.4M & \textbf{0.666} & \textbf{0.800}\\
        \bottomrule
    \end{tabular}}
\end{table}

%%%%%%% [E] VTAB Results  %%%%%%%%%%
\section{VTAB preliminary results}
\label{suppsec:vtab}
We also perform a small evaluation over a subset of the popular Visual Task Adaptation Benchmark (VTAB), using an ImageNet-21k pretrained ViT-B.
As can be seen, \textsl{ETHER} and \textsl{ETHER+} perform comparably to OFT with $n=256$ and LoRA rank $8$, while using a fraction of the trainable parameters.
\vspace{-0.3cm}

\begin{table}[H]
    \centering
    \caption{VTAB results}
    \label{tab:vtab}
    \vspace{0.1cm}
    \scalebox{0.93}{
    \begin{tabular}{l|c|cccc|c|c}
        \toprule
        & \multirow{2}{*}{\#params}& \multicolumn{4}{c|}{\textbf{Natural}} & \textbf{Specialized} & \textbf{Structured} \\ 
        & & Caltech101&DTD&Flowers102&SVHN&  EuroSAT & sNORB-Elev\\ 
        \midrule
        Full Finetuning &	85.8M&	96.26&73.03&98.71&73.71&	96.16&	63.36\\
        Linear Probing&	0&	95.96&72.34&\textbf{99.12}&52.55&	95.03&	34.09\\
        LoRA$_{r=8}$&	1.33M&	97.69&\textbf{77.50}&99.10&\textbf{97.40}&	98.92&	74.89\\
        OFT$_{n=256}$&	0.29M&	96.95&75.80&98.60&96.58&	98.83&	74.37\\
        \textsl{ETHER}&	0.08M&97.64&75.85&98.83&95.81&	98.80&	74.17\\
        \textsl{ETHER+}&	0.33M	&\textbf{98.27}&76.92&98.88&96.84&	\textbf{99.15}&	\textbf{78.41}\\
        \bottomrule
    \end{tabular}}
\end{table}

\end{document}